\documentclass[letterpaper, 10 pt, journal, twoside]{IEEEtran}
\usepackage{amsmath,amsfonts}
\usepackage{algorithmic}
\usepackage{algorithm}
\usepackage{array}
\usepackage[caption=true,font=footnotesize,labelfont=rm,textfont=rm]{subfig}
\usepackage{textcomp}
\usepackage{stfloats}
\usepackage{url}
\usepackage{verbatim}
\usepackage{graphicx}
\usepackage{cite}
\usepackage{ulem}

\usepackage{bbding}
\usepackage{bm}
\usepackage{multirow}
\usepackage[Symbol]{upgreek}
\usepackage{amsfonts}
\usepackage{diagbox}
\usepackage{amsmath}

\usepackage{caption}
\usepackage{subcaption}
\usepackage{pifont}
\usepackage{booktabs}
\usepackage{tabularx}

\usepackage{newtxtext} 
\usepackage{newtxmath} 
\usepackage{graphicx} 
\usepackage{subcaption} 
\usepackage{multirow}
\usepackage{cite}
\usepackage[]{hyperref}

\begin{document}
\title{LXD-SLAM: LiDAR+X Dense SLAM with $\sum_{i=0}^{5}C_5^i$ Configurable Sensor Combinations}
\author{
Zhong~Wang$^{1}$, Lin~Zhang$^{*1}$, Linfei Li$^1$, Ying Shen$^{1}$, Shaoming Zhang$^2$, Pengcheng Shi$^3$, and Shengjie~Zhao$^{1}$
\thanks{This work was supported in part by the National Natural Science Foundation of China under Grant 62403308 (\textit{Corresponding author: Lin Zhang})}
\thanks{$^{1}$Zhong Wang, Lin Zhang, Linfei Li, Ying Shen, and Shengjie Zhao are with the School of Computer Science, Tongji University, Shanghai, 201804, China (email: \{cszhongwang, cslinzhang, 2311444, yingshen,  zhaoshengjie\}@tongji.edu.cn).}
\thanks{$^{2}$Shaoming Zhang is with the School of Surveying and Geo-Informatics, Tongji University, Shanghai, 200092, China (email:zhangshaoming@tongji.edu.cn).}
\thanks{$^{3}$Pengcheng Shi is with School of Remote Sensing and Information Engineering, Wuhan University, Wuhan, 430072, China (email: shipc\_2021@whu.edu.cn).}
}

\markboth{Manuscript Submitted to XXX.}
{Wang \MakeLowercase{\textit{et al.}}: LXD-SLAM: LiDAR+X Dense SLAM with $\sum_{i=0}^{5}C_5^i$ Configurable Sensor Combinations} 
\maketitle

\begin{abstract}
Simultaneous Localization and Mapping (SLAM) is essential for autonomous systems, yet achieving reliable, globally consistent pose estimation and dense mapping in complex environments remains challenging due to geometric degeneracy and sensor drift. While multi-sensor fusion addresses these issues, existing systems often lack the modularity to adapt to diverse platforms and rely on mathematically inconsistent fusion or suboptimal map representations. To address these limitations, we propose LXD-SLAM (LiDAR+X Dense SLAM), a highly versatile and unified multi-sensor fusion framework. Centered around 3D LiDAR, our system allows for the plug-and-play integration of LiDAR, Camera, IMU, Wheel Encoder, and GNSS, supporting up to 32 distinct sensor combinations. We employ a mathematically unified Iterative Error-Sate Kalman Filter with an adaptive hierarchical prediction strategy and an update step that minimizes point-to-mesh distances and visual reprojection errors. To support this, the environment is modeled using continuous multi-layered Gaussian Process (GP) sub-meshes, which enables efficient ray-to-mesh depth recovery for visual features. For global consistency, we introduce an Extended Scan Context (ESC) descriptor derived from the GP sub-meshes alongside a Bidirectional PnP optimization for robust multi-modal loop closure within a hybrid pose graph. Extensive evaluations on public datasets and real-world experiments demonstrate that LXD-SLAM matches or exceeds state-of-the-art specialized odometry solutions across various configurations while generating high-fidelity, globally consistent dense meshes in real-time. The relevant codes and data will be made available at \url{https://github.com/peterWon/LXD-SLAM} upon publication.
\end{abstract}

\begin{IEEEkeywords}
LiDAR SLAM, Dense Meshing
\end{IEEEkeywords}

\section{Introduction}
Simultaneous Localization and Estimation (SLAM) is a fundamental capability for autonomous systems, enabling them to operate in unknown environments. In recent years, 3D Light Detection and Ranging (LiDAR) sensors have become a cornerstone of robust SLAM due to their ability to provide accurate, long-range measurements independent of ambient lighting. This has led to their widespread adoption in diverse fields, from autonomous driving and service robotics to large-scale surveying and mapping. However, achieving reliable and globally consistent pose estimation and dense mapping in complex, large-scale environments remains a significant challenge. The inherent sparsity of LiDAR data in certain directions, susceptibility to geometric degeneracy (e.g., long tunnels or open spaces), and sensor drift over time necessitate the fusion of complementary sensors~\cite{10286080}.

To this end, recent years have witnessed a substantial amount of research on LiDAR-based multi-sensor fusion SLAM. In such area, a prevalent direction is LiDAR-inertial odometry (LIO)~\cite{ye2019tightly,shan2020lio,he2020lioekf,qin2020lins,xu2022fastlio2,cao2022fasterlio}, which tightly couples high-frequency IMU data with LiDAR point clouds to smooth motion estimation and compensate for motion distortion; To further enhance robustness in visually rich but geometrically degenerate scenes, LiDAR-visual or LiDAR-visual-inertial odometry (LVIO) systems~\cite{zhang2015visual,shan2021lvi,zuo2019lic,lin2021r2live,lin2022r3live,zheng2022fastlivo,fast-livo2} have sprung up, fusing visual features to provide additional constraints; For scenarios demanding maximum coverage and redundancy, multi-LiDAR systems have been developed to enhance field of view and point density~\cite{jiao2021robust}; For ground-moving platforms, fusion with wheel encoders offers a complementary odometry source that is particularly robust in geometrically featureless environments~\cite{yuan2023liwo}; Beyond odometry, achieving global consistency requires the integration of absolute or relative loop closure measurements. This demand prompts LiDAR-GNSS fusion for absolute pose corrections in open-sky areas by coupling GNSS signals~\cite{shan2020lio,qin2019vins,cao2022gvins}.

Despite these pleasing advances, existing systems often suffer from the following limitations. Most of them are designed for a fixed sensor suite, lacking the modularity to adapt to different platforms or mission requirements. 
Their fusion architectures can be mathematically inconsistent, loosely coupling different sensor modalities, which can lead to suboptimal accuracy, especially under challenging conditions. Furthermore, their representations of the environment are often bottlenecks: sparse feature-based maps discard valuable geometric information, while globally dense reconstructions are computationally intractable for large-scale environments and often do not provide a continuous, smooth surface model suitable for direct and accurate data association.

To solve the above-mentioned problem, driven by the need for a more versatile, dense, and globally consistent solution, we propose a unified multi-sensor fusion SLAM framework with LiDAR as its core, supporting additional LiDAR (L), Camera (C), IMU (I), Wheel Encoder (W), and GNSS (G) inputs. 
Let the set of supported modalities be: $$\mathcal{S}=\{\textbf{L}, \textbf{C}, \textbf{I}, \textbf{W}, \textbf{G}\}.$$
This \textbf{L}iDAR+\textbf{X} \textbf{D}ense \textbf{SLAM} system, \textbf{LXD-SLAM} for short, is configurable for up to $\sum_{i=0}^{5}C_5^i=32$ different types of multi-sensor fusion with: 
$$\textbf{X}\in\mathcal{P}(\mathcal{S}),$$ 
where $\mathcal{P}$($\mathcal{S}$) denotes the power set of $\mathcal{S}$.

When designing LXD-SLAM, our primary motivation is to create a unified but modular framework that can seamlessly accommodate an arbitrary combination of sensors while maintaining a high level of estimation accuracy and building a globally consistent dense map.
The core working principle of our system is as follows. 
The environment is represented as a collection of local sub-meshes. For each LiDAR scan registered to a sub-mesh, the points are discretized into grid cells, and a multi-layered Gaussian Process (GP) surface is fitted to the points within each cell, capturing complex geometries with up to three orientated surface layers. 
This continuous representation directly supplies point-to-mesh constraints for the front-end. 
For state estimation, we employ an Iterative Error-State Kalman Filter (IESKF) with a hierarchical prediction strategy: prioritized IMU propagation, followed by wheel-encoder-based dead reckoning, and finally a constant-velocity model with rotation estimation strategy, depending on the sensors available. 
The filter's update step primarily minimizes point-to-mesh errors. 
When camera data is available, visual features are tracked via optical flow, their depths are recovered through ray tracing against the mesh and fused using a 1D inverse depth filter, and the reprojection errors of mature features provide additional update signals. 
In the back-end, a hybrid pose graph of keyframes and sub-mesh poses is maintained, and loop closures from LiDAR, vision, or GNSS are fused to achieve globally consistent trajectory and map optimization. This tightly integrated design forms the basis of our main contributions:

\begin{enumerate}
    \item \textbf{Unrivaled Sensor Versatility and Plug-and-Play Fusion:} We present, to the best of our knowledge, the first SLAM system supporting up to $2^5$ distinct sensor combinations. By designating LiDAR as the primary sensor, the framework allows for the arbitrary, plug-and-play integration of additional LiDAR, camera, IMU, wheel encoder, and GNSS. This extreme modularity enables the system to be effortlessly tailored to a wide variety of robotic platforms and mission-specific requirements without fundamental re-engineering.
    \item \textbf{A Unified Estimation Framework with Adaptive Prediction and Update}: We advocate a mathematically unified registration and fusion architecture based on IESKF. The filter state is adaptively configured to include pose, velocity, and, when available, IMU biases. A hierarchical strategy is proposed to support state prediction of the system under different internal perception sensor configurations. The state is updated by minimizing a primary point-to-mesh distance error, with systematic deviations further corrected by jointly optimizing visual reprojection errors from processed visual features. This ensures stable and accurate state estimation across varying sensor availability.
    \item \textbf{Robust and Efficient Visual Feature Processing with Ray-to-Mesh Depth Association}: We develop a robust visual feature manager that employs optical flow for efficient cross-frame feature tracking. For each tracked feature, a hierarchical ray tracing algorithm against the reconstructed mesh is introduced to efficiently recover its absolute depth. This depth is then fused over multiple frames within a sliding window using a 1D inverse depth filter, ensuring that only mature, reliably estimated features contribute to the system's state update. Descriptors are extracted for these mature features, and together with their associated depths, which are lifted to 3D keypoints and utilized for loop closure detection and constraint construction in the back-end.
    \item \textbf{Globally Consistent Mapping via Multi-modal Loop Closure and GP-derived Topology}: We propose a robust backend framework that ensures global consistency by leveraging the rich geometric and visual information inherent in our GP-based representation.
    First, we introduce an Extended Scan Context (ESC) descriptor that maps the multi-layered GP sub-meshes into a three-channel polar representation—encoding height, surface density, and visibility. This allows for drift-free topological loop detection that is resilient to dynamic occlusions and sparse sensing. Second, we develop a Bidirectional PnP optimization for visual loop closure. Unlike standard methods, it treats the historical map and current query with geometric symmetry, establishing a unified relative pose constraint from non-reciprocal 3D-2D correspondences. By integrating these constraints into a hybrid factor graph alongside GNSS and odometry, our system achieves high-fidelity, globally consistent dense reconstruction in complex, large-scale environments.
\end{enumerate}

The proposed system is rigorously evaluated across a wide range of public datasets and real-world experiments. The results demonstrate that our method achieves performance comparable to or exceeding specialized SOTA solutions across all supported sensor configurations. Furthermore, we showcase the system's ability to maintain high-fidelity, globally consistent dense meshes in real-time, providing a robust foundation for downstream autonomous tasks.

\section{Related Work}
Simultaneous Localization and Mapping (SLAM) has evolved significantly from its single-sensor origins to embrace multi-modal fusion, driven by the complementary nature of different sensors. This section reviews existing literature across the key fusion paradigms relevant to our work.

\subsubsection{LiDAR-Inertial SLAM/Odometry}
The integration of an Inertial Measurement Unit (IMU) with LiDAR significantly mitigates motion distortion and improves high-frequency pose tracking. Early pioneering works like LOAM \cite{zhang2014loam}, LeGO-LOAM \cite{shan2018legoloam}, and F-LOAM \cite{wang2021floam} utilized loose coupling, extracting sparse edge and planar features for frame-to-frame or frame-to-map registration. To improve robustness in challenging environments, tightly coupled frameworks emerged, which are broadly categorized into optimization-based and filter-based methods. 

Optimization-based approaches formulate state estimation over a sliding window or factor graph. LIOM \cite{ye2019tightly} pioneered tight coupling by jointly optimizing LiDAR features and IMU pre-integration. Building on this, LIO-SAM \cite{shan2020lio} employs a factor graph optimization backend to incorporate marginalization and various sensor factors, while LiO-EKF \cite{he2020lioekf} presents an alternative coupled formulation. Conversely, filter-based methods, primarily built upon the Error-State Kalman Filter (ESKF), offer remarkable computational efficiency. Early implementations like LINS \cite{qin2020lins} utilized a robot-centric formulation. Modern benchmarks like FAST-LIO \cite{xu2021fastlio} and its successor FAST-LIO2 \cite{xu2022fastlio2} utilize an Iterated Kalman Filter paired with an incremental kd-tree (\textit{ikd-tree}), directly registering raw points without explicit feature extraction. More recently, frameworks like Point-LIO \cite{bai2022pointlio} and Faster-LIO \cite{cao2022fasterlio} have extended these concepts to handle extreme field-of-view, non-repetitive scanning patterns, and faster voxels. Despite their success, these LIO systems primarily rely on sparse features or discrete point clouds for mapping, which fail to capture the continuous geometric structure of the environment and lack the modularity to easily incorporate additional sensor modalities.

\subsubsection{LiDAR-Visual-Inertial SLAM/Odometry}
To overcome the geometric degeneracy of LiDAR in structureless environments (e.g., long corridors) and visual failures in low-texture settings, LiDAR-Visual-Inertial fusion has attracted extensive research. VLOAM \cite{zhang2015visual} and DEMO \cite{ji2006demo} demonstrated early success by loosely coupling visual odometry with LiDAR scan matching. Recent state-of-the-art systems utilize tight coupling to fully exploit cross-modal correlations. Among optimization-based systems, LVI-SAM \cite{shan2021lvi} builds upon LIO-SAM, running parallel visual-inertial (VIO) and LiDAR-inertial (LIO) subsystems that assist each other via graph optimization. LIC-fusion \cite{zuo2019lic} and LIC-fusion 2.0 \cite{zuo2020lic2} introduce online spatiotemporal calibration within a sliding window framework. 

On the filter-based side, R$^2$LIVE \cite{lin2021r2live} and R$^3$LIVE \cite{lin2022r3live} employ an IESKF to fuse camera, LiDAR, and IMU data, with the latter reconstructing a real-time radiance map. Similarly, FAST-LIVO \cite{zheng2022fastlivo} and its outstanding successor Fast-LIVO2 \cite{fast-livo2} build upon the IESKF framework by directly minimizing the photometric errors of map patches, bypassing traditional feature matching. However, associating accurate depths to visual features remains a key challenge; many systems rely on crude point-cloud projection (e.g., VINS-Mono with LiDAR depth \cite{qin2018vinsmono}), which is highly susceptible to occlusion, sparsity, and parallax errors. In contrast, our framework leverages a reconstructed dense mesh and ray-tracing to robustly extract absolute depth for visual features, feeding a 1D inverse depth filter for highly stable tracking and update.

\subsubsection{Multi-LiDAR SLAM}
Expanding the field of view (FoV) and point density through multiple LiDARs is a direct approach to enhancing spatial awareness and robustness against degeneracy in large-scale applications. Early efforts focused heavily on offline extrinsic calibration. For online operation, M-LOAM \cite{jiao2021robust} proposes a robust system for multi-LiDAR real-time calibration and mapping, efficiently managing feature extraction across multiple heterogeneous sensors. Similarly, systems designed for autonomous driving often utilize multiple hardware-synchronized LiDARs to eliminate blind spots, as prominently featured in large-scale benchmarks like Waymo \cite{sun2020scalability}, nuScenes \cite{caesar2020nuscenes}, and KITTI-360 \cite{liao2022kitti360}.

To handle multi-LiDAR state estimation, some frameworks employ decentralized or centralized multi-sensor graph structures \cite{shang2026mlsslam}. More recent real-time systems, such as MLA-OM \cite{song2024mliomab} and Multi-LIO \cite{li2024multifast}, attempt to parallelize point cloud processing from different heads. However, existing Multi-LiDAR systems are usually heavily engineered for specific, fixed sensor arrays and demand strict extrinsic priors. They lack a true plug-and-play architecture. Our framework natively designates a primary LiDAR while treating additional LiDARs as supplementary update sources within the unified IESKF, allowing for arbitrary Multi-LiDAR configurations without altering the core mathematical structure.

\subsubsection{LiDAR-Wheel and LiDAR-GNSS Fusion}
For ground vehicles and large-scale outdoor mapping, wheel encoders and GNSS are indispensable sources of bounded or absolute positioning. Wheel encoders provide highly reliable dead reckoning in feature-deprived indoor environments (e.g., smooth tunnels) where both LiDAR and vision might suffer from geometric or photometric degeneracy \cite{wu2017vins}. Incorporating wheel kinematics via standard odometry models (e.g., differential-drive or Ackerman models) has been widely adopted in ground robot SLAM \cite{yuan2023liwo}. GNSS, on the other hand, eliminates accumulated drift in long-range outdoor trajectories by providing global position priors. 
Systems like LIO-SAM \cite{shan2020lio}, VINS-Fusion \cite{qin2019vins} incorporate GNSS measurements as global factors in their factor graph optimization backends. To tightly integrate these modalities, frameworks like GVINS \cite{cao2022gvins} directly fuse raw GNSS pseudoranges into a visual-inertial system, while LIWO \cite{yuan2023liwo} tightly couples LiDAR, inertial, and wheel odometry. 
However, these systems typically handle these sensors in an ad-hoc or loosely coupled manner designed for specific vehicle kinematics, risking system failure if a primary sensor drops out.

\subsubsection{Dense Map-Based LiDAR SLAM}
The representation of the map fundamentally dictates the accuracy of data association, loop closure detection, and the utility of the SLAM output for downstream tasks. Traditional point-cloud maps are memory-intensive and discrete, lacking topological continuity. Surfel-based methods, such as SuMa \cite{behley2018efficient} and SuMa++ \cite{chen2019suma}, project point clouds into 2.5D range images to build dense surfel maps, but this projection inherently loses 3D geometric information in complex multi-layer structures. Voxel-based methods like VoxelMap \cite{yuan2022voxelmap} and BALM \cite{liu2021balm} model the environment probabilistically by fitting planes or distributions within voxels. While efficient, a single plane per voxel cannot adequately represent complex, intersecting geometries (e.g., corners, foliage). 

Recently, neural implicit representations (e.g., NeRF-based and 3D Gaussian Splatting SLAM) \cite{zhu2022nice, matsuki2024gaussian} have provided continuous, photo-realistic mapping but suffer from severe computational overhead and catastrophic forgetting, rendering them unsuitable for real-time iterative Kalman filter front-ends. Traditional continuous surface representations using Gaussian Processes (GP) \cite{ruan2020gpslam} have also been explored to model implicit continuous occupancy or elevation surfaces, but they traditionally face cubic computational complexity $O(N^3)$ relative to point size. Our approach significantly advances this domain by leveraging multi-layered GP sub-meshes. By adaptively allowing up to three oriented surface layers within each grid cell, we achieve a mathematically continuous, memory-efficient, and highly accurate dense representation that natively provides distance-to-mesh constraints for the front-end and supports global consistency in the back-end.

\section{Methodology}
\begin{figure*}
    \centering
    \includegraphics[width=\linewidth]{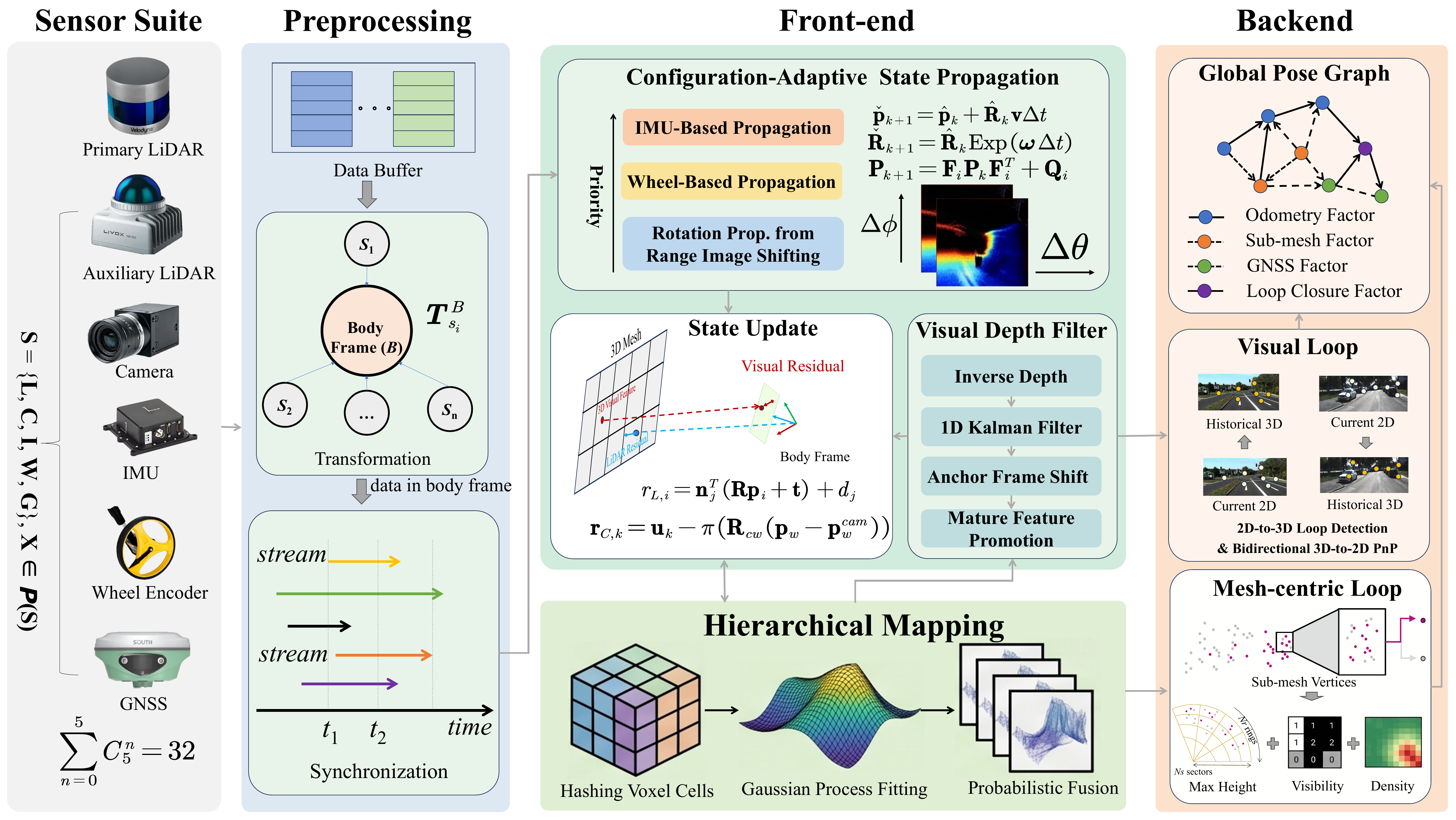}
    \caption{System Overview. The proposed LXD-SLAM framework infrastructure is anchored by a primary LiDAR and architected to support the tight-coupled fusion of heterogeneous proprioceptive and exteroceptive sensors. Temporal synchronization aligns multi-modal streams in the asynchronous preprocessing stage. The frontend tracks the continuous state updates via a generalized Itered Error-State Kalman Filter utilizing geometric and optional visual residuals. Long-term drift is suppressed globally in the backend via a non-linear factor graph optimization utilizing multi-source topological and structural constraints.}
    \label{fig:sys}
\end{figure*}

\subsection{System Overview}
As illustrated in Fig. \ref{fig:sys}, the proposed LXD-SLAM framework establishes a primary LiDAR as the structural anchoring reference and features a highly modular topology designed to seamlessly ingest arbitrary combinations of multi-modal streams from the sensor set $\mathcal{S}=\{\textbf{L}, \textbf{C}, \textbf{I}, \textbf{W}, \textbf{G}\}$. This generalized architectural configuration dynamically instantiates itself into one of $\sum_{i=0}^{5}C_5^i=32$ operational sensor-fusion configurations. 

To resolve asynchronous sampling, the preprocessing layer enforces rigorous spatiotemporal alignment by chronologically registering heterogeneous sensor data relative to the master LiDAR clock while concurrently performing spatial extrinsic transformations to maintain a unified ego-motion frame. In the frontend tracking module, state estimation is governed by a unified Iterative Error-State Kalman Filter (IESKF) framed as a localized manifold optimization. The filter propagates state kinematics via a hierarchical backup strategy—prioritizing high-frequency inertial odometry and systematically degenerating to wheel encoder dead-reckoning or kinematic constant-velocity models depending on the instantaneous structural survival of the sensor suite. State updates are subsequently driven by the simultaneous minimization of geometric point-to-mesh residuals and optional multi-view visual reprojection constraints. To guarantee global consistency and bound cumulative drift, the backend maintains a sparse factor graph that executes batch optimization over heterogeneous constraints, including relative odometry factors, structural node-to-mesh topology factors, absolute GNSS position priors, and dual-domain loop closures verified via mesh-centric geometric descriptors and symmetric bidirectional perspective-n-point alignments.

\subsection{Star-Topology Extrinsic Registration}
In complex multi-sensor systems, the structural representation of spatial transformations (extrinsics) dictates both the estimation accuracy and the systemic robustness. Traditional robotic architectures typically model kinematic configurations as a sequential tree topology ($\text{LiDAR} \rightarrow \text{Camera} \rightarrow \text{IMU} \rightarrow \text{Body}$). Although structurally intuitive, such chained topologies introduce geometric bottlenecks when handling dynamic hardware configurations. To circumvent these failures, we advocate a centralized, dynamic two-layer star topology for multi-sensor extrinsic management.

\subsubsection{Dynamic Base Frame Arbitrage}
In contradistinction to conventional algorithms that bind state variables to a statically defined base link, our framework dynamically designates the global ``Body" frame $B$ conditioned on the active sensor topology verified during initialization. The base frame allocation is governed by a deterministic kinematic priority sequence: $\text{IMU} > \text{Wheel Encoder} > \text{Camera} > \text{Primary LiDAR}$. By binding the coordinate origin to the highest-frequency proprioceptive sensor available, the system optimizes the mathematical stability of the IESKF kinematic propagation step under high-dynamic motion profiles.

\subsubsection{Star-Topology Mapping Formulation} Let $\mathbf{T}_{S_i}^{B} \in SE(3)$ denote the rigid body transformation from the $i$-th active sensor frame $S_i$ to the dynamically arbitrated Body frame $B$:
\begin{equation}
    \mathbf{T}_{S_i}^{B} = \begin{bmatrix} \mathbf{R}_{S_i}^{B} & \mathbf{t}_{S_i}^{B} \\ \mathbf{0}^T & 1 \end{bmatrix},
\end{equation}
where $\mathbf{R}_{S_i}^{B} \in SO(3)$ and $\mathbf{t}_{S_i}^{B} \in \mathbb{R}^3$. Rather than maintaining a complex graph of relative transformations between adjacent nodes, the configuration workspace projects all active sensor systems directly onto the central node $B$, rendering a flat, two-layer star network. This architectural paradigm provides two fundamental advantages for resilient multi-sensor localization:

\begin{itemize}
    \item {Decoupling and Elimination of Cascading Calibration Noise}: In a conventional chain topology, extracting the spatial relationship of a distal sensor requires compound matrix multiplications along the kinematic path:
    \begin{equation}
        \mathbf{T}_{S_n}^{B} = \mathbf{T}_{S_1}^{B} \mathbf{T}_{S_2}^{S_1} \dots \mathbf{T}_{S_n}^{S_{n-1}}.
    \end{equation}
    Under this formulation, localized calibration perturbations $\delta \mathbf{T}$ or numerical inaccuracies within intermediate nodes compound multiplicatively, yielding severe lever-arm errors. By contrast, the star topology guarantees $\mathcal{O}(1)$ direct mathematical access to $\mathbf{T}_{S_i}^{B}$, completely shielding the system from cascading transformation errors and ensuring strict geometric consistency across the entire multi-sensor cluster.
    \item {Graph Completeness and Plug-and-Play Resilience}: To accommodate up to 32 distinct operational sensor permutations without structural modification, the underlying spatial graph must resist node deletion. Within a tree structure, the sudden failure or deliberate removal of an intermediate sensor breaks the transformation chain, isolating all downstream nodes. The star configuration structurally resolves this dependency: because each sensor retains an independent, unchained mathematical link to the core Body frame, the runtime activation or suspension of any sensor subset operates entirely in isolation, maintaining the mathematical integrity of the remaining active sub-graph.
\end{itemize}

\subsection{Temporal Synchronization and Preprocessing}
To preserve the mathematical assumptions of the frontend estimator, incoming data streams undergo rigid chronological alignment. This synchronization layer structurally transforms asynchronous, multi-rate sensor inputs—including raw LiDAR point clouds, high-frequency IMU vectors, and visual frames—into a coherent temporal sequence.

\subsubsection{Multi-Rate Data Synchronization}
For high-frequency proprioceptive streams (e.g., IMU), a circular buffer architecture isolates the temporal measurements spanning consecutive LiDAR frame captures, providing the foundational discrete inputs for kinematic state propagation. For lower-frequency or intermittent absolute observations (e.g., GNSS), temporal mapping matches external timestamps to the closest internal state epoch within a bounded, configurable optimization gate, ensuring that exteroceptive residuals are projected onto valid temporal states.

\subsubsection{Kinematic Motion Distortion Compensation}
Mechanical scanning LiDARs introduce structural distortion due to the ego-motion of the vehicle during the continuous acquisition of a single scan sweep. Utilizing precise per-point acquisition timestamps $\Delta t_i$, raw point clouds are rigidly deskewed to the terminal epoch of the scan sweep. By querying the predicted continuous state trajectories generated by the frontend tracking filter, the relative transformation from the point sampling instance to the scan end time is resolved. Rotational components are regularized using Spherical Linear Interpolation (SLERP), whereas translational components follow linear interpolation, projecting the warped spatial observations strictly onto a single time-invariant reference frame.

\subsection{Hierarchical Scene Representation and Management}
\begin{figure}[t]
    \centering
    \includegraphics[width=\linewidth]{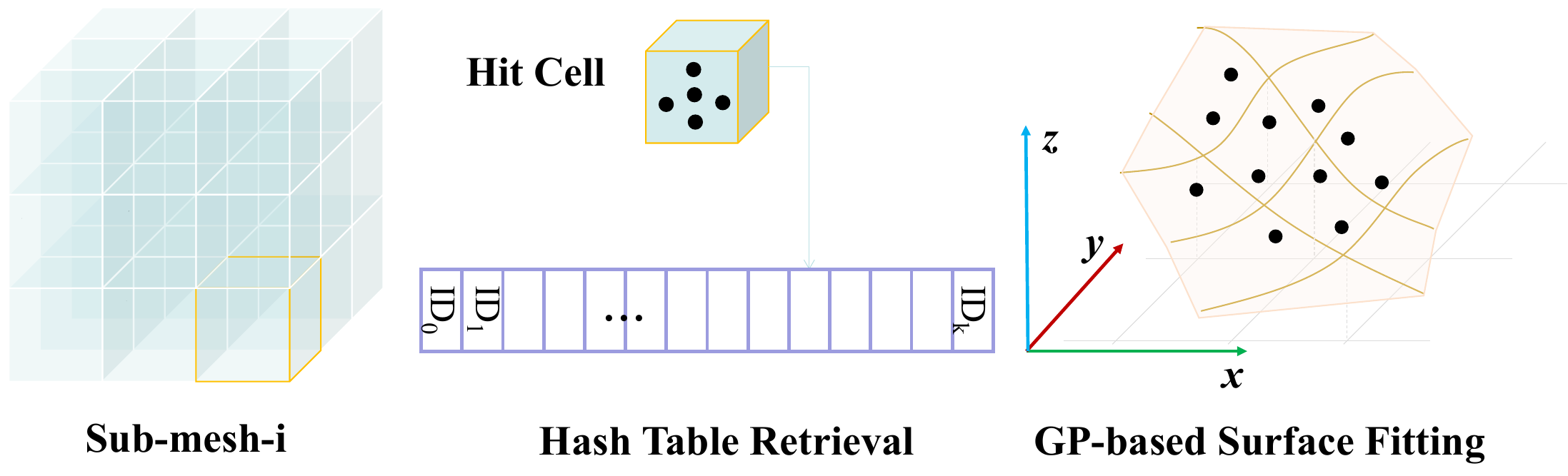}
    \caption{Hierarchical map organization. The continuous 3D workspace is uniformly partitioned into discrete cells and indexed via spatial hashing. Within each valid cell, continuous surface reconstruction is performed through local Gaussian Process regression driven by structural dimensionality assessments.}
    \label{fig:scene-manage}
\end{figure}

To facilitate real-time data association and high-frequency state updates, inspired by SLAMesh~\cite{10161425}, we model the spatial surrounding via a disciplined structural hierarchy: a global topological map composed of sliding-window local sub-meshes, which are further discretized into uniformly distributed volumetric cells, with each cell indexed by a hash table and mesh fitted by Gaussian Process regression as Fig. \ref{fig:scene-manage} illustrates.

\subsubsection{Bounded Computational Complexity Sub-Mesh Management}
To prevent the computational degradation associated with scaling monolithic geometric maps, the environment is discretized into a chronological sequence of localized sub-meshes managed by an active submap ring-buffer. Each sub-mesh is bound to encapsulate a finite number of sequential frames. At any execution instance, only the immediate temporal sub-meshes are retained within the active tracking pipeline for point cloud registration and probabilistic surface refinement. This formulation caps the computational footprint of spatial maintenance with respect to total trajectory length, ensuring real-time optimization invariance.

\subsubsection{Spatial Hashing and Dynamic Cell Allocation}
Within each localized sub-mesh, the continuous 3D Euclidean space is partitioned into cubic cells of a fixed resolution $r_{\text{cell}}$. To enforce memory efficiency and guarantee constant-time queries ($\mathcal{O}(1)$), active cells are allocated dynamically via a spatial hash map. A 3D coordinate vector $\mathbf{p} = [x, y, z]^T \in \mathbb{R}^3$ is deterministically mapped to a discrete voxel index vector $\mathbf{v} = [v_x, v_y, v_z]^T \in \mathbb{Z}^3$:
\begin{equation}
    \mathbf{v} = \left\lfloor \frac{\mathbf{p}}{r_{\text{cell}}} \right\rfloor.
\end{equation}
The coordinate index $\mathbf{v}$ is subsequently compressed into a unique scalar hash key through optimized bitwise operations. This hash-driven spatial organization bypasses the overhead of conventional hierarchical tree structures, enabling instantaneous spatial neighbor discovery and accelerated ray-casting pipelines.

\subsubsection{Continuous Surface Reconstruction via Gaussian Process Regression}
In contrast to standard Truncated Signed Distance Fields (TSDF) that necessitate dense, computationally burdensome volumetric integrations, our approach reconstructs mathematically continuous manifold surfaces directly from sparse observations via localized Gaussian Process (GP) regression.

Prior to executing the Gaussian Process inference, a structural dimensionality assessment is conducted using Principal Component Analysis (PCA) on the raw points accumulated within a target cell to evaluate the local geometric topology. Let $\lambda_1 \le \lambda_2 \le \lambda_3$ represent the sorted eigenvalues derived from the local point covariance matrix. A cell is classified as non-structural (e.g., vegetation, transient noise) and pruned from the mapping pipeline if the structural ratio of the dominant eigenvalues violates a geometric planarity threshold:
\begin{equation}
    \frac{\lambda_3}{\lambda_2} > \tau_{ns}.
\end{equation}
Conversely, if the cell is validated as a structurally consistent surface, the computed eigenvectors define an optimal projection plane. The complex 3D surface reconstruction problem is thus structurally reduced to a localized 2D-to-1D depth regression mapping, significantly decreasing the computational complexity of the subsequent inference step.

Let the downsampled training points within a validated cell be parameterized within the localized projection plane as $\mathbf{X}$ with corresponding depth observations $\mathbf{y}$:
\begin{equation}
    \mathbf{X} = \{ \mathbf{x}_1, \dots, \mathbf{x}_N \} \in \mathbb{R}^{2 \times N}, \quad \mathbf{y} = \{ y_1, \dots, y_N \}^T \in \mathbb{R}^N.
\end{equation}
We formulate the underlying continuous surface model as a Gaussian Process:
\begin{equation}
    f(\mathbf{x}) \sim \mathcal{GP}\left(m(\mathbf{x}), k(\mathbf{x}, \mathbf{x}')\right).
\end{equation}
An isotropic exponential covariance kernel is employed to model the spatial correlation between distinct coordinates:
\begin{equation}
    k(\mathbf{x}_i, \mathbf{x}_j) = \exp \left( -l \left\| \mathbf{x}_i - \mathbf{x}_j \right\|_2 \right),
\end{equation}
where $l$ defines the characteristic length-scale parameter. Given a set of uniformly distributed target evaluation coordinates $\mathbf{X}^* \in \mathbb{R}^{2 \times M}$ representing the desired mesh vertices, the predictive mean vector $\mathbf{f}^*$ and its associated predictive covariance matrix $\mathbf{\Sigma}^*$ are evaluated analytically via:
\begin{align}
    \mathbf{f}^* &= \mathbf{K}_{*N} (\mathbf{K}_{NN} + \sigma_n^2 \mathbf{I})^{-1} \mathbf{y}, \\
    \mathbf{\Sigma}^* &= \mathbf{K}_{**} - \mathbf{K}_{*N} (\mathbf{K}_{NN} + \sigma_n^2 \mathbf{I})^{-1} \mathbf{K}_{N*},
\end{align}
where $\mathbf{K}_{NN}$ denotes the $N \times N$ cross-covariance matrix of the training set, $\mathbf{K}_{*N} = \mathbf{K}_{N*}^T$ defines the covariance matrix between target evaluation points and training inputs, and $\sigma_n^2$ models the forward sensor measurement noise variance.

As the vehicle traverses the environment, cells receive recurring temporal observations. The system recursively fuses newly inferred GP surfaces with historical map structures using an inverse-variance weighting scheme. For a target vertex possessing a prior historical mean $\mu_{\text{old}}$ and variance $\sigma_{\text{old}}^2$, and an incoming independent observation $\mu_{\text{obs}}$ with variance $\sigma_{\text{obs}}^2$, the optimal probabilistic fusion is given by:
\begin{equation}
    \mu_{\text{new}} = \frac{\mu_{\text{old}}\sigma_{\text{obs}}^2 + \mu_{\text{obs}}\sigma_{\text{old}}^2}{\sigma_{\text{old}}^2 + \sigma_{\text{obs}}^2}, \quad \sigma_{\text{new}}^2 = \frac{\sigma_{\text{old}}^2 \sigma_{\text{obs}}^2}{\sigma_{\text{old}}^2 + \sigma_{\text{obs}}^2}.
\end{equation}
This formulation yields an inherently smooth, mathematically continuous manifold map that naturally filters dynamic outliers and sensor noise.

\subsection{Frontend: Unified State Estimation}
The tracking backend of the frontend architecture utilizes an Iterated Error-State Kalman Filter (IESKF)~\cite{ieskf} to tightly fuse multi-modal sensor configurations. The estimator is explicitly formulated to adapt its kinematic models and observational constraints at runtime based on the functional survival of the sensor suite.

\subsubsection{Generalized Filtering Framework and Error-State Formulation}
To mathematically handle the non-linearities inherent to the $3D$ manifold space $SO(3)$, the state tracking problem is cast as a recursive localized optimization over an 18-dimensional error state vector defined on the tangent space. At any time step $k$, the true state vector is decomposed into a nominal state vector $\mathbf{x}_k$ and an error state vector $\delta\mathbf{x}_k$ through a generalized manifold compound operator $\boxplus$. The continuous-time nominal state vector is defined as:
\begin{equation}
    \mathbf{x} = [\mathbf{R}^T, \mathbf{p}^T, \mathbf{v}^T, \mathbf{b}_g^T, \mathbf{b}_a^T, \mathbf{g}^T]^T,
\end{equation}
where $\mathbf{R} \in SO(3)$ maps the orientation of the body frame into the global world reference, $\mathbf{p}, \mathbf{v} \in \mathbb{R}^3$ define the position and linear velocity vectors in the world frame, $\mathbf{b}_g, \mathbf{b}_a \in \mathbb{R}^3$ denote the gyroscope and accelerometer biases, and $\mathbf{g} \in \mathbb{R}^3$ represents the localized gravity vector. The associated 18-dimensional tangent error state is parameterized as:
\begin{equation}
    \delta\mathbf{x} = [\delta\boldsymbol{\theta}^T, \delta\mathbf{p}^T, \delta\mathbf{v}^T, \delta\mathbf{b}_g^T, \delta\mathbf{b}_a^T, \delta\mathbf{g}^T]^T \in \mathbb{R}^{18}.
\end{equation}
Given a prior state estimate $\check{\mathbf{x}}_k$ with its associated covariance matrix $\mathbf{P}_k$, and a multi-modal measurement pool $\mathcal{M}$ containing geometric and visual observations, the IESKF update phase acts as a localized Maximum A Posteriori (MAP) optimizer. It solves for the optimal error state increment $\delta\mathbf{x}$ at the $\kappa$-th iteration by minimizing the joint cost function:
\begin{equation}
    \delta \mathbf{x}^{\kappa+1} = \arg\min_{\delta\mathbf{x}} \left( \|\delta\mathbf{x} - \delta\mathbf{x}^\kappa\|_{\mathbf{P}_k^{-1}}^2 + \sum_{m \in \mathcal{M}} \|\mathbf{r}_m(\hat{\mathbf{x}}_k^\kappa \boxplus \delta\mathbf{x})\|_{\boldsymbol{\Sigma}_m^{-1}}^2 \right),
    \label{eq:ieskf_framework}
\end{equation}
where $\hat{\mathbf{x}}_k^\kappa$ represents the nominal state evaluated at the current iteration step, $\mathbf{r}_m(\cdot)$ denotes the generalized sensor residual function, and $\boldsymbol{\Sigma}_m$ represents the observation noise covariance. Upon convergence, the nominal state vector is updated via the composition mapping $\mathbf{R} \leftarrow \mathbf{R} \exp(\delta\boldsymbol{\theta}^\wedge)$ and standard vector addition for Euclidean components, after which the error state vector is reset to zero.

\subsubsection{Configuration-Adaptive Kinematic State Propagation}
Based on the active sensor topology identified by the centralized manager, the system dynamically alters the state transition equations used to compute the prior state $\check{\mathbf{x}}_{k+1}$ and propagate the covariance matrix $\mathbf{P}_{k+1}$.

\textit{IMU-Driven Inertial Propagation}: When an active IMU is available, high-frequency inertial measurements drive the nominal system state transition equations. Given raw gyroscope observations $\boldsymbol{\omega}_m$ and accelerometer observations $\mathbf{a}_m$ at discrete interval $k$, the nominal state propagates forward via:
\begin{align}
    \check{\mathbf{R}}_{k+1} &= \hat{\mathbf{R}}_k \text{Exp}\left((\boldsymbol{\omega}_m - \hat{\mathbf{b}}_{g,k})\Delta t\right), \\
    \check{\mathbf{v}}_{k+1} &= \hat{\mathbf{v}}_k + \left(\hat{\mathbf{R}}_k(\mathbf{a}_m - \hat{\mathbf{b}}_{a,k}) + \hat{\mathbf{g}}_k\right)\Delta t, \\
    \check{\mathbf{p}}_{k+1} &= \hat{\mathbf{p}}_k + \hat{\mathbf{v}}_k \Delta t + \frac{1}{2}\left(\hat{\mathbf{R}}_k(\mathbf{a}_m - \hat{\mathbf{b}}_{a,k}) + \hat{\mathbf{g}}_k\right)\Delta t^2,
\end{align}
where $\text{Exp}(\cdot)$ maps an element from the tangent space $\mathfrak{so}(3)$ to the Lie group $SO(3)$. The state error covariance matrix $\mathbf{P}$ is projected forward via linearization:
\begin{equation}
    \mathbf{P}_{k+1} = \mathbf{F}_i \mathbf{P}_k \mathbf{F}_i^T + \mathbf{Q}_i,
\end{equation}
where $\mathbf{F}_i \in \mathbb{R}^{18 \times 18}$ represents the continuous-time error state transition matrix evaluated as:
\begin{equation}
    \mathbf{F}_i = \begin{bmatrix} 
    \text{Exp}(-\hat{\boldsymbol{\omega}}\Delta t) & \mathbf{0} & \mathbf{0} & -\mathbf{J}_r \Delta t & \mathbf{0} & \mathbf{0} \\
    \mathbf{0} & \mathbf{I} & \mathbf{I}\Delta t & \mathbf{0} & \mathbf{0} & \mathbf{0} \\
    -\hat{\mathbf{R}}_k(\mathbf{a}_m - \hat{\mathbf{b}}_a)^\wedge \Delta t & \mathbf{0} & \mathbf{I} & \mathbf{0} & -\hat{\mathbf{R}}_k \Delta t & \mathbf{I}\Delta t \\
    \mathbf{0} & \mathbf{0} & \mathbf{0} & \mathbf{I} & \mathbf{0} & \mathbf{0} \\
    \mathbf{0} & \mathbf{0} & \mathbf{0} & \mathbf{0} & \mathbf{I} & \mathbf{0} \\
    \mathbf{0} & \mathbf{0} & \mathbf{0} & \mathbf{0} & \mathbf{0} & \mathbf{I}
    \end{bmatrix},
\end{equation}
where $\mathbf{J}_r$ defines the right Jacobian of $SO(3)$, and $\mathbf{Q}_i$ models the integrated covariance of the sensor measurement noise and random-walk biases.

\textit{Wheel-Odometry Kinematic Propagation}: In scenarios where the IMU stream suffers terminal degradation, state propagation defaults to a chassis kinematics model driven by wheel encoders and a derived base angular velocity vector $\boldsymbol{\omega}$. The forward translation vector is resolved by projecting the body-frame odometry speed $\mathbf{v}_{\text{wheel}}$ onto the world reference frame:
\begin{equation}
    \check{\mathbf{p}}_{k+1} = \hat{\mathbf{p}}_k + \hat{\mathbf{R}}_k \mathbf{v}_{\text{wheel}} \Delta t, \quad \check{\mathbf{R}}_{k+1} = \hat{\mathbf{R}}_k \text{Exp}(\boldsymbol{\omega} \Delta t).
\end{equation}
Covariance propagation is updated via a specialized wheel-odometry transition structure:
\begin{equation}
    \mathbf{P}_{k+1} = \mathbf{F}_w \mathbf{P}_k \mathbf{F}_w^T + \mathbf{Q}_w \Delta t,
\end{equation}
where $\mathbf{F}_w$ is a decoupled sub-matrix of $\mathbf{F}_i$ that isolates the acceleration-bias constraints, and $\mathbf{Q}_w$ explicitly models the non-holonomic slippage uncertainties of the mobile platform.

\textit{Range Image Shift Orientation Propagation}: In highly degraded configurations where both IMU and wheel encoders are absent, the system infers egocentric rotation vectors from sequential raw point cloud alignments using range image shifts. A single-frame raw 3D point cloud $\mathcal{P} = \{\mathbf{p}_i = (x_i, y_i, z_i) \in \mathbb{R}^3\}_{i=1}^N$ is projected onto a 2D cylindrical range grid $\mathcal{I} \in \mathbb{R}^{H \times W}$, where $H$ and $W$ represent the vertical and horizontal angular sampling parameters. For any valid point $\mathbf{p}_i$, its range distance $d_i$, yaw coordinate $\theta$, and pitch coordinate $\phi$ are evaluated via:
\begin{equation}
    d_i = \sqrt{x_i^2 + y_i^2 + z_i^2}, \quad \theta = \arctan2(y_i, x_i), \quad \phi = \arcsin\left(\frac{z_i}{d_i}\right).
\end{equation}
These polar values map directly to discrete image pixel indices $(u, v)$ under the boundary conditions of the sensor field-of-view ($\phi_{\text{up}}, \phi_{\text{down}}$):
\begin{equation}
    u = \left\lfloor \left(0.5 - \frac{\theta}{2\pi}\right) \cdot W \right\rceil, \quad v = \left\lfloor \frac{\phi_{\text{up}} - \phi}{\phi_{\text{up}} - \phi_{\text{down}}} \cdot H \right\rceil.
\end{equation}
Spatial occlusions are resolved by implementing a Z-buffer sequence that retains only the minimum radial range coordinate:
\begin{equation}
    \mathcal{I}(u, v) = \min_{\mathbf{p}_k \mapsto (u,v)} d_k.
\end{equation}
Because the horizontal and vertical pixel dimensions of $\mathcal{I}$ map linearly to spherical coordinates, structural column and row shifts $\mathbf{s} = [\Delta u, \Delta v]^T$ between consecutive sweeps isolate the internal angular increments ($\Delta \theta, \Delta \phi$):
\begin{equation}
    \Delta \theta = -\frac{\Delta u}{W} \cdot 2\pi, \quad \Delta \phi = \frac{\Delta v}{H} \cdot (\phi_{\text{up}} - \phi_{\text{down}}).
\end{equation}
Setting the pitch variation to zero ($\Delta \psi = 0$) based on non-holonomic ground vehicle constraints, the angular increments are mapped to a rotation matrix $\mathbf{R} \in SO(3)$ using a Z-Y-X quaternion composition sequence:
\begin{equation}
    \mathbf{q} = \mathbf{q}_z(\Delta \theta) \otimes \mathbf{q}_y(\Delta \phi) \otimes \mathbf{q}_x(0).
\end{equation}
The derived rotation vector isolates the angular velocity parameters, allowing linear velocities to be generated through a constant-velocity kinematic assumption, which standardizes the covariance propagation equations according to the wheel-odometry formulation.

\subsubsection{Multi-Modal Manifold Observation Constraints}
Once a synchronized LiDAR sweep or visual frame is registered by the estimator, the joint cost function described in Eq. \eqref{eq:ieskf_framework} is instantiated using specific operational observation residuals.

\textit{LiDAR Point-to-Mesh Structural Constraints}: For a deskewed LiDAR point observation $\mathbf{p}_i$ expressed within the body coordinate reference frame, the system queries the active local sub-mesh to isolate the nearest surface cell structure. The geometric residual $\mathbf{r}_L$ is formulated as the orthogonal point-to-mesh distance:
\begin{equation}
    r_{L, i} = \mathbf{n}_j^T (\mathbf{R}\mathbf{p}_i + \mathbf{p}) + d_j,
\end{equation}
where $\mathbf{n}_j$ and $d_j$ define the normal vector and distance parameters of the planar vertex regression. The analytical Jacobian matrix $\mathbf{H}_{L, i}$ with respect to the tangent state error components $[\delta\boldsymbol{\theta}^T, \delta\mathbf{p}^T]^T$ is derived as:
\begin{equation}
    \mathbf{H}_{L, i} = [\mathbf{n}_j^T (\mathbf{R}\mathbf{p}_i)^\wedge \quad \mathbf{n}_j^T].
\end{equation}
Crucially, the information weight assigned to this residual is scaled by the inverse of the localized Gaussian Process variance $\sigma_{\text{mesh}}^2$ associated with that specific mesh vertex. This formulation seamlessly injects map uncertainty bounds directly into the filter updates.

\textit{Visual Reprojection Constraints}: For a monitored visual feature bound to a 3D landmark coordinate $\mathbf{p}_w$, the observation residual $\mathbf{r}_C$ is mapped onto the normalized image plane:
\begin{equation}
    \mathbf{r}_{C, k} = \mathbf{u}_{k} - \pi\left(\mathbf{R}_{CB} (\mathbf{R}^T (\mathbf{p}_w - \mathbf{p}) - \mathbf{t}_{CB})\right),
\end{equation}
where $\mathbf{u}_k$ denotes the normalized coordinate observation, $\pi(\cdot)$ represents the standard pinhole perspective projection model, and $\mathbf{R}_{CB}, \mathbf{t}_{CB}$ represent the static extrinsic transformation from the camera frame to the body center.

\subsubsection{Visual Feature Lifecycle and Depth Filtering}
\begin{figure}[t]
    \centering
    \includegraphics[width=0.9\linewidth]{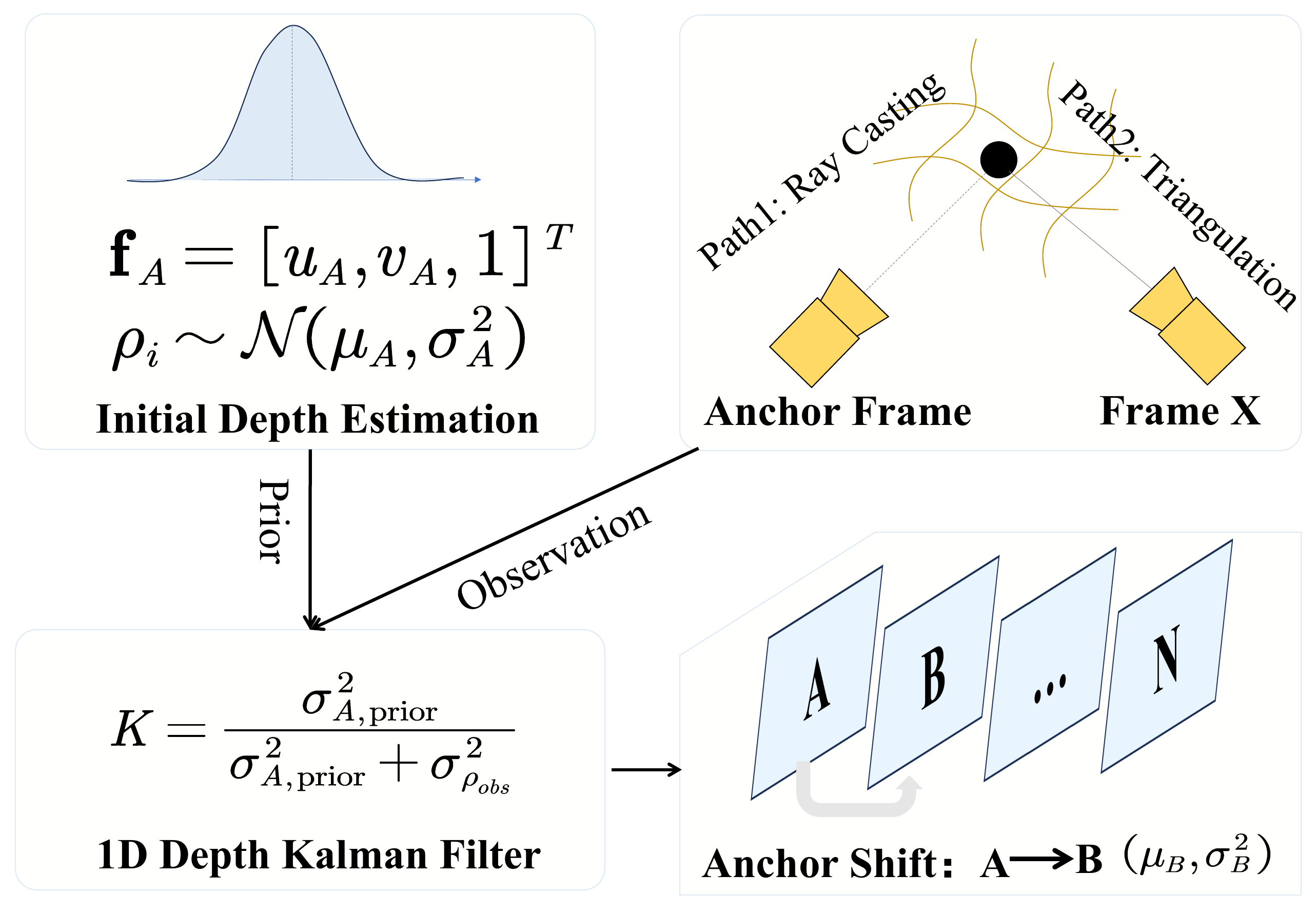}
    \caption{Visual feature lifecycle topology. Features are tracked across sliding-window frames via inverse depth parameterization anchored to their frame of origin. Tracking continuity is maintained across frame marginalization through error-propagated anchor shifts.}
    \label{fig:feature_manage}
\end{figure}

To maintain tracking continuity during multi-sensor tracking, visual features are monitored via an inverse depth filter operating on a sliding window which is shown in Fig. \ref{fig:feature_manage}.

To model uncertainty fields across arbitrary depths, each tracked feature is parameterized by its inverse depth $\rho$ within the camera reference frame of its initialization, designated as the anchor frame $A$. For a specific feature $i$, its state distribution is modeled as a Gaussian density: $\rho_i \sim \mathcal{N}(\mu_A, \sigma_A^2)$. The feature's position is initialized as a normalized coordinate vector $\mathbf{f}_A = [u_A, v_A, 1]^T$. When an observation is registered within a subsequent frame $C$, the relative transformation between coordinates is defined as $\mathbf{T}_{CA} = [\mathbf{R}_{CA} \mid \mathbf{t}_{CA}]$.

If a valid range observation $d_c$ is verified at the feature coordinate via mesh ray-casting, the 3D point projects onto the anchor frame to resolve the analytical depth value $d_A$, yielding an inverse depth observation vector $\rho_{\text{obs}} = 1/d_A$. The associated observation noise variance is approximated through first-order propagation:
\begin{equation}
    \sigma_{\rho_{\text{obs}}}^2 = \frac{\sigma_{d_A}^2}{d_A^4}.
\end{equation}

In the absence of direct geometric depth intersections, depth estimation defaults to multi-view triangulation. The system computes the parallax angle $\alpha$ across observation rays. If the angular separation satisfies baseline requirements, triangulation is performed using a Direct Linear Transformation (DLT). The uncertainty of this triangulated estimate is modeled as a function of baseline length $\|\mathbf{t}_{CA}\|$ and parallax geometry, yielding the measurement variance $\sigma_{\rho_{\text{obs}}}^2$. Both observational modes are recursively integrated into the feature prior using a 1D Kalman Filter update:
\begin{equation}
    K = \frac{\sigma_{A, \text{prior}}^2}{\sigma_{A, \text{prior}}^2 + \sigma_{\rho_{\text{obs}}}^2},
\end{equation}
\begin{equation}
    \mu_{A, \text{post}} = \mu_{A, \text{prior}} + K (\rho_{\text{obs}} - \mu_{A, \text{prior}}),
\end{equation}
\begin{equation}
    \sigma_{A, \text{post}}^2 = (1 - K) \sigma_{A, \text{prior}}^2.
\end{equation}

To enforce strict boundary limits on computational overhead, the active tracking frame pool is constrained to a maximum size $N$. When the frame capacity is saturated, the oldest optimization frame is marginalized. If the marginalized frame is identified as the active anchor $A$ for a feature track, an anchor shift is executed to transfer the state history to the next valid frame $B$ within the sliding window. Let the relative transformation between these frames be given by $\mathbf{T}_{BA} = [\mathbf{R}_{BA} \mid \mathbf{t}_{BA}]$, where $\mathbf{r}_3^T$ represents the terminal row vector of $\mathbf{R}_{BA}$ and $t_z$ denotes the vertical translation component. To preserve geometric consistency, the transformed inverse depth mean $\mu_B$ is evaluated via:
\begin{equation}
    \mu_B = \frac{\mu_A}{\mathbf{r}_3^T \mathbf{f}_A + \mu_A t_z}.
\end{equation}
Defining the scalar shorthand $C = \mathbf{r}_3^T \mathbf{f}_A$, the analytical Jacobian mapping $J = \frac{\partial \mu_B}{\partial \mu_A}$ is resolved via:
\begin{equation}
    J = C \left( \frac{\mu_B}{\mu_A} \right)^2.
\end{equation}
The propagated state variance within the new anchor frame is updated via $\sigma_B^2 = J^2 \sigma_A^2$, and the normalized feature vector is re-projected onto the coordinate frame of frame $B$. A feature track is promoted to the global graph backend if it remains tracked across a minimum of three sequential frames and its inverse depth variance satisfies a convergence threshold ($\sigma^2 < 0.05$).

\subsubsection{Mesh-Assisted Visual Depth Association}
To correlate high-fidelity geometric depth bounds with sparse visual points, the tracking frontend implements a mesh-assisted spatial intersection algorithm. This approach leverages the continuous surface representations within the active sub-meshes to execute ray-triangle intersection tests.

For each tracked visual feature, the system generates a unit ray vector $\mathbf{d}_f$ within the sub-mesh frame. By traversing the voxelized cell network using a 3D Digital Differential Analyzer (DDA)~\cite{3d-dda} pipeline, potential intersecting surface regions are discovered without exhausting the global point pool. Once an intersection with a fitted surface mesh triangle is validated, the 3D position vector $\mathbf{p}_f$ and its associated Gaussian Process variance metric $\sigma_f^2$ are extracted.

The integration pipeline operates through three hierarchical operations. A feature coordinate $\mathbf{u}$ is back-projected into a normalized unit vector using calibrated camera intrinsics. This vector is subsequently transformed into the local sub-mesh reference using the current predicted state to formulate an active search ray $\mathbf{r}(t) = \mathbf{o}_s + t\mathbf{d}_s$. To optimize computational throughput, the DDA pipeline determines cell entry and exit bounds across the spatial hash map, bypassing empty space and restricting intersection routines to voxels that contain validated Gaussian Process surfaces. Within a candidate cell, the ray is tested against the local mesh triangles constructed from the GP predictive means. Upon a confirmed intersection, barycentric interpolation resolves the 3D position vector, and the local Gaussian Process variance $\sigma_f^2$ is mapped directly to the visual feature. This variance operates as a dynamic weight during the joint IESKF optimization, de-weighting visual updates originating from sparse or highly uncertain map segments.

\subsection{Backend: Robust Loop Closure and Joint Optimization}
To enforce global structural consistency and bound cumulative drift fields across extended trajectories, the backend architecture executes a global pose graph optimization. The non-linear factor graph integrates relative odometry inputs, localized sub-mesh constraints, LiDAR loop factor entries, multi-view visual constraints, and intermittent absolute GNSS priors.

\subsubsection{Global Factor Graph Formulation}
The backend optimization problem is cast as a non-linear batch optimization over a sparse graph network. Let $\mathcal{T}_b = \{\mathbf{T}_{WB_0}, \mathbf{T}_{WB_1}, \dots, \mathbf{T}_{WB_n}\}$ define the set of historical vehicle body frames, and $\mathcal{T}_m = \{\mathbf{T}_{WM_0}, \mathbf{T}_{WM_1}, \dots, \mathbf{T}_{WM_m}\}$ represent the set of sub-mesh poses, where $\mathbf{T}_{WB_i}, \mathbf{T}_{WM_j} \in SE(3)$. The optimization objective solves for the optimal states $\mathcal{T}_b^*$ and $\mathcal{T}_m^*$ that minimize the joint Mahalanobis distance across all graph constraints:
\begin{equation}
    \mathcal{T}_b^*, \mathcal{T}_m^* = \arg\min_{\mathcal{T}_b, \mathcal{T}_m} \sum_{k \in \mathcal{F}} \rho \left( \| \mathbf{r}_k \|_{\mathbf{\Sigma}_k}^2 \right),
\end{equation}
where $\mathcal{F}$ represents the comprehensive set of active measurement factors, $\mathbf{\Sigma}_k$ is the covariance matrix associated with the $k$-th factor, and $\rho(\cdot)$ represents a Huber robust cost function configured to mitigate outliers caused by perceptual aliasing or multi-path GNSS reception. The structural objective function isolates four distinct cost factors:

\begin{itemize}
    \item {Relative Odometry Factors}: Constraints between consecutive vehicle frames derived from frontend tracking updates:
    \begin{equation}
        \mathbf{r}_{\text{odom}}(i, i+1) = \log \left( \Delta \tilde{\mathbf{T}}_{i, i+1}^{-1} (\mathbf{T}_{WB_i}^{-1} \mathbf{T}_{WB_{i+1}}) \right)^\vee.
    \end{equation}
    \item {Absolute GNSS Factors}: Intermittent global position constraints anchored to the localized Cartesian East-North-Up (ENU) origin:
    \begin{equation}
        \mathbf{r}_{\text{gnss}}(i) = \text{Trans}(\mathbf{T}_{WB_i}) - \tilde{\mathbf{p}}_{\text{gnss}, i},
    \end{equation}
    where $\text{Trans}(\cdot)$ extracts the 3D translation components of the transformation matrix.
    \item {Sub-Mesh Local Factors}: Relative constraints anchoring historical vehicle frames to their corresponding active sub-mesh centers to preserve local dense surface consistency:
    \begin{equation}
        \mathbf{r}_{\text{mesh}}(i, j) = \log \left( \Delta \tilde{\mathbf{T}}_{i, j}^{-1} (\mathbf{T}_{WB_i}^{-1} \mathbf{T}_{WM_j}) \right)^\vee.
    \end{equation}
    \item {Loop Closure Factors}: Relative pose constraints $\Delta \tilde{\mathbf{T}}_{p, c}$ established between a historical keyframe $\mathbf{T}_{WB_p}$ and a current frame $\mathbf{T}_{WB_c}$, instantiated via dual-modality loop detection modules.
\end{itemize}

\subsubsection{Visual Loop Closure via Bidirectional PnP Realignment}
Conventional visual loop closure systems enforce strict 3D-3D or 3D-2D feature correspondences. However, in long-term operations, triangulated landmarks across distant temporal frames frequently lack direct 3D structural overlap due to viewpoint variations and occlusion fields. To leverage visual feature inputs without enforcing restrictive bidirectionality, we propose a Bidirectional Perspective-and-Point (PnP) formulation. This approach decouples feature matching into two independent sets of 3D-2D projection constraints while optimizing a single relative transformation matrix $\mathbf{T}_{CH} \in SE(3)$ mapping from the historical frame reference $\{H\}$ to the current frame reference $\{C\}$.

Let $\mathcal{A}$ define the set of forward matches where historical 3D landmarks $\mathbf{P}_{H, i}$ project onto current 2D normalized coordinates $\mathbf{p}_{C, i}$. Let $\mathcal{B}$ represent the set of backward matches where current 3D points $\mathbf{P}_{C, j}$ map onto historical 2D coordinates $\mathbf{p}_{H, j}$. The projection residual errors for a candidate rotation $\mathbf{R}_{CH}$ and translation $\mathbf{t}_{CH}$ are defined as:
\begin{align}
    \mathbf{e}_{A, i} &= \mathbf{p}_{C, i} - \pi(\mathbf{R}_{CH} \mathbf{P}_{H, i} + \mathbf{t}_{CH}), \\
    \mathbf{e}_{B, j} &= \mathbf{p}_{H, j} - \pi(\mathbf{R}_{CH}^{\top} (\mathbf{P}_{C, j} - \mathbf{t}_{CH})),
\end{align}
where $\pi(\cdot)$ maps a 3D coordinate to its normalized image position. Robust initial pose hypotheses are generated by running a standard P3P RANSAC pipeline~\cite{ransac} over the forward set $\mathcal{A}$. Candidate configurations are subsequently verified by evaluating total inlier counts across both verification sets $\mathcal{A}$ and $\mathcal{B}$. Finally, the relative transformation is refined through joint non-linear optimization over all validated inliers:
\begin{equation}
    \mathbf{T}_{CH}^{*} = \arg\min_{\mathbf{T}_{CH}} \left( \sum_{i \in \mathcal{A}} \rho(||\mathbf{e}_{A, i}||_{\Sigma}^{2}) + \sum_{j \in \mathcal{B}} \rho(||\mathbf{e}_{B, j}||_{\Sigma}^{2}) \right),
\end{equation}
and the optimal relative pose factor is integrated as a loop edge within the global factor graph.

\subsubsection{Mesh-Centric Loop Closure via Extended Scan Context}
To provide geometric redundancy in texture-degraded or variable-illumination environments, the backend extracts global topological descriptors directly from the reconstructed sub-meshes. The sub-meshes provide continuous surface manifold profiles bounded by explicit Gaussian Process variance estimates. We construct an Extended Scan Context (ESC) descriptor by projecting trusted mesh vertices onto a polar grid defined by $N_r$ rings and $N_s$ sectors centered around the sensor origin. To filter out low-confidence structural structures, a vertex is pruned if its associated GP variance exceeds a reliable threshold: $\sigma^2 > \tau_{var}$. For a valid vertex $\mathbf{P}_{S, k}$ in the sub-mesh coordinate reference, the point is transformed into the sensor's local coordinate frame to derive its polar coordinates $(\rho, \theta)$, which isolate its cell index within the ESC matrix.

\textit{Confidence-Aware Density Channel via GP Variance}: In conventional point-cloud global descriptors, spatial density channels are constructed by counting raw point returns within each polar sector bin. However, this counting mechanism becomes ill-posed when applied to a continuous, uniformly sampled sub-mesh architecture. Because local surfaces within each cell are reconstructed and resampled via Gaussian Process regression, counting extracted vertices merely reflects the surface area of the local mesh, completely discarding the underlying sensor observation density and measurement uncertainty.

To resolve this limitation, we reformulate the density channel by weighting point inputs with their predictive GP variance. The variance parameter encodes local structural quality, capturing both the density of the original raw sensor observations and the geometric consistency of the underlying surface. We propose a confidence-aware density channel where the cell density metric $D(r, \theta)$ aggregates the inverse GP variance:
\begin{equation}
    D(r, \theta) = \sum_{k \in \mathcal{S}_{r, \theta}} \frac{1}{\sigma_k^2 + \epsilon},
\end{equation}
where $\mathcal{S}_{r, \theta}$ represents the set of projected surface vertices falling within polar bin $(r, \theta)$, $\sigma_k^2$ is the predicted GP variance for the $k$-th vertex, and $\epsilon$ is a small positive regularization constant. This formulation ensures that stable, structured geometric landmarks (e.g., planar building facades) exhibit low variance and dominate the density matrix. Conversely, unstructured, transient, or noisy elements (e.g., vegetation, dynamic obstacles) yield high variance, and their influence is naturally suppressed. This eliminates the need for auxiliary segmentation modules.

\textit{FoV-Agnostic Visibility Channel via 2.5D Raycasting}: Standard global descriptors assume a full $360^{\circ}$ panoramic field-of-view (FoV). However, practical deployments utilizing narrow-FoV sensors or navigating highly cluttered environments often produce spatially truncated or severely occluded local submaps. Directly comparing partially observable descriptors using standard distance metrics introduces a significant penalty bias, leading to high false-negative rates.

To address this issue, we augment the descriptor with a visibility channel $\mathcal{V}(r, \theta)$ that encodes the deterministic occlusion state of the local layout. Instead of employing expensive 3D volumetric ray-tracing, a lightweight 2.5D polar raycasting operation is executed. For a given azimuth sector $\theta$, the visibility state along the radial direction $r$ is defined as:
\begin{equation}
    \mathcal{V}(r, \theta) = \begin{cases} 
    1 \ (\textit{free}), & \text{if } r < r_{\text{occ}}(\theta) \\
    2 \ (\textit{occupied}), & \text{if } r = r_{\text{occ}}(\theta) \\
    0 \ (\textit{unknown}), & \text{if } r > r_{\text{occ}}(\theta) \lor \textit{unobserved}
    \end{cases}
\end{equation}
where $r_{\text{occ}}(\theta)$ represents the radial distance to the first encountered surface vertex within sector $\theta$. This visibility channel operates as a dynamic mask during the descriptor matching phase. When computing the column-wise shift similarity between a query descriptor $\mathcal{V}^q$ and a historical candidate $\mathcal{V}^c$, the distance metric is exclusively evaluated on sectors where both descriptors exhibit valid mutually observable states ($\mathcal{V} \neq 0$). By masking out unknown and unobserved regions, the similarity evaluation becomes invariant to sensor FoV discrepancies and viewpoint-dependent occlusions, ensuring robust loop detection in constrained environments.

\section{Experiment}
\subsection{Setup}
\subsubsection{Implementation}

All experiments were conducted on a laptop equipped with an Intel Core i9-14900HX CPU and 32 GB of RAM. All system modules were implemented in C++. The Robot Operating System (ROS) was utilized to interface with various sensors and manage the publishing and subscribing of the corresponding data messages. Furthermore, the system relies on OpenCV for image processing and employs GTSAM to construct and optimize the global pose graph.
\subsubsection{Dataset}
To verify the effectiveness of our system, we conduct experiments on three widely adopted and representative multi-sensor datasets: NTU-VIRAL, FusionPortableV2, and KITTI. In addition, we collect a campus dataset using a self-built device to evaluate the system's real-world performance.

\textbf{NTU-VIRAL}~\cite{viral}. It is designed for UAV platforms and was collected around typical buildings at NTU, covering diverse and complex UAV motions. It is equipped with two OS1-16 (Gen 1) LiDARs (one oriented horizontally and one vertically), two uEye 1221 LE cameras, and a 9-axis IMU (VectorNav VN100). This dataset provides precise time synchronization between the LiDARs and cameras, establishing a solid foundation for multi-sensor fusion. It offers ground-truth positions (acquired by a 3D Laser Tracker, Leica MS60 TotalStation), enabling accurate quantitative evaluation of the poses estimated by SLAM.

\textbf{FusionPortableV2}~\cite{fusionportablev2}. The sensors in FusionPortableV2 are integrated into a portable kit that can be conveniently deployed on different carriers. Its data collection platforms cover common carriers targeted by SLAM, including handheld, quadruped robot dog, UGV, and vehicle. The sensor suite consists of a 128-beam Ouster OS1 LiDAR with a maximum range of 120 m, a pair of FILR BFS-U3-31S4C stereo cameras with a resolution of 1024×768, and a 6-axis STIM300 IMU. In some outdoor scenarios, GNSS measurements are also provided, and certain UGV sequences provide wheel odometry measurements. The dataset supplies 6DoF ground-truth poses obtained from surveying-grade instruments (3DM-GQ7, Leica MS60, Leica BLK360, and RTC360).

\textbf{KITTI}~\cite{KITTI}. KITTI is a widely used dataset for testing autonomous driving algorithms. It includes stereo cameras and a Velodyne 64-beam LiDAR. The dataset covers a wide range of environments, including complex road conditions such as urban streets and mountain roads. Some sequences feature long durations and abundant loop closures, allowing effective evaluation of the system's globally consistent mapping capability.

\begin{figure}
    \centering
    \includegraphics[width=\linewidth]{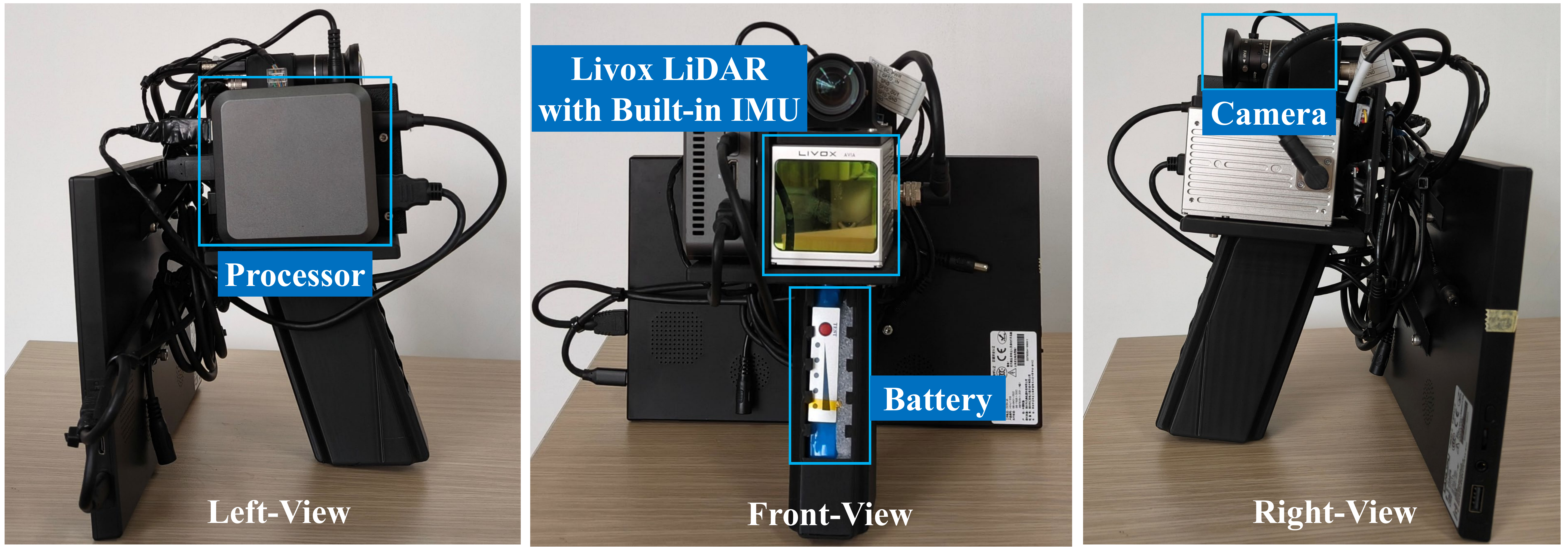}
    \caption{Self-developed Device.}
    \label{fig:device}
\end{figure}

\textbf{Self-collected dataset.} We acquire multi-sensor data in campus environments using a handheld device as illustrated in Fig. \ref{fig:device}. The device is equipped with a Livox LiDAR, an industrial camera, and a built-in 6-axis IMU. The point cloud and image data are synchronized via a microcontroller to ensure accurate fusion of raw measurements. We collect several sequences in both indoor and outdoor settings: indoor sequences are captured with the handheld device, while outdoor sequences are collected by holding the device while moving on an electric bicycle.

\subsection{Results of Multi-sensor Odometry}
To validate the plug-and-play versatility and estimation accuracy of the proposed LXD-SLAM across diverse sensor suites, we conduct extensive evaluations on the NTU-VIRAL~\cite{viral} and FusionPortableV2~\cite{fusionportablev2} datasets. For a fair and rigorous comparison, all estimated trajectories are aligned with the ground truth using the Umeyama algorithm, and the absolute trajectory error (ATE) Root Mean Square Error (RMSE) is adopted as the primary metric for localization fidelity.

\begin{table*}[htbp]
    \centering
    \caption{RMSEs of competing odometry systems on NTU-VIRAL (m). Results that diverge or are with RMSEs larger than 2m are denoted by \ding{55}.}
    \label{tab:rmse-viral}
    \scriptsize
    \setlength{\tabcolsep}{2pt}
    \begin{tabular}{l|cc|cc|ccccccc|ccccc}
    \hline
    \hline
    \multirow{2}{*}{\textbf{Dataset}} & \multicolumn{2}{c|}{\textbf{LiDAR-Camera}} & \multicolumn{2}{c|}{\textbf{LiDAR-LiDAR}} & \multicolumn{7}{c|}{\textbf{LiDAR-IMU}} & \multicolumn{5}{c}{\textbf{LiDAR-IMU-Camera}} \\
    & \textbf{SDV-LOAM} & \textbf{Ours(lc)} & \textbf{MLOAM} & \textbf{Ours(ll)} & \textbf{D-LIOM} & \textbf{Ct-LIO} & \textbf{dlio} & \textbf{LIO-SAM} & \textbf{Fast-LIO2} & \textbf{KN-LIO} & \textbf{Ours} & \textbf{R$^3$LIVE} & \textbf{LVI-SAM} & \textbf{Fast-LIVO} & \textbf{Fast-LIVO2} & \textbf{Ours} \\
    \midrule
    eee01 & \textbf{0.301} & \ding{55} & \textbf{0.249} & \ding{55} & 0.191 & 0.186 & 0.143 & 0.074 & 0.068 & 0.072 & \textbf{0.038} & 0.072 & 3.901 & 0.191 & 0.068 & \textbf{0.034} \\
    eee02 & \ding{55} & \textbf{0.066} & 0.166 & \textbf{0.051} & 0.508 & 0.416 & 0.141 & 0.067 & 0.075 & 0.071 & \textbf{0.029} & 0.059 & 0.182 & 0.132 & 0.051 & \textbf{0.029} \\
    eee03 & 0.301 & \textbf{0.102} & \textbf{0.232} & 2.269 & 0.220 & 0.326 & 0.186 & 0.117 & 0.109 & 0.087 & \textbf{0.034} & 0.078 & 0.287 & 0.192 & 0.068 & \textbf{0.033} \\
    nya01 & 0.202 & \textbf{0.054} & 0.123 & \textbf{0.052} & 0.154 & 0.160 & 0.110 & 0.077 & 0.060 & 0.075 & \textbf{0.035} & 0.080 & 0.205 & 0.121 & 0.073 & \textbf{0.033} \\
    nya02 & 0.214 & \textbf{0.073} & 0.191 & \textbf{0.064} & 0.200 & 0.159 & 0.154 & 0.091 & 0.095 & 0.083 & \textbf{0.034} & 0.084 & 1.296 & 0.182 & 0.075 & \textbf{0.030} \\
    nya03 & 0.251 & \textbf{0.070} & 0.226 & \textbf{0.062} & 0.202 & 0.203 & 0.175 & 0.147 & 0.103 & 0.083 & \textbf{0.031} & 0.079 & 0.176 & 0.112 & 0.059 & \textbf{0.030} \\
    sbs01 & 0.212 & \textbf{0.062} & 0.173 & \textbf{0.058} & 0.178 & 0.165 & 0.146 & 0.088 & {0.084} & 0.083 & \textbf{0.038} & 0.075 & 0.254 & 0.253 & 0.062 & \textbf{0.037} \\
    sbs02 & 0.233 & \textbf{0.072} & 0.147 & \textbf{0.057} & 0.165 & 0.176 & 0.133 & {0.089} & 0.076 & 0.091 & \textbf{0.045} & 0.076 & 0.221 & 0.134 & 0.061 & \textbf{0.042} \\
    sbs03 & 0.281 & \textbf{0.060} & 0.151 & \textbf{0.058} & 0.154 & 0.375 & 0.146 & 0.083 & 0.076 & 0.080 & \textbf{0.039} & 0.070 & 0.309 & 0.132 & 0.060 & \textbf{0.037} \\
    \hline
    \hline
    \end{tabular}
\end{table*}

\begin{table*}[htbp]
    \centering
    \caption{RMSEs of competing odometry systems on FusionPortableV2 (m). Results that diverge or are with RMSEs larger than 2m are denoted by \ding{55}.}
    \label{tab:rmse-fpv2}
    \scriptsize
    \setlength{\tabcolsep}{4pt}
    \begin{tabular}{l|c|c|cc|ccccc|cccc}
    \hline
    \hline
    \multirow{2}{*}{} & \multicolumn{1}{c|}{\textbf{LiDAR}} & \multicolumn{1}{c|}{\textbf{LiDAR-Camera}} & \multicolumn{2}{c|}{\textbf{LiDAR-IMU-Wheel}} & \multicolumn{5}{c|}{\textbf{LiDAR-IMU}} & \multicolumn{4}{c}{\textbf{LiDAR-IMU-Camera}} \\
    \textbf{Dataset} & \textbf{Ours} & \textbf{Ours (lc)} & \textbf{LIWO} & \textbf{Ours (liw)} & \textbf{dlio} & \textbf{Fast-LIO2} & \textbf{Faster-LIO} & \textbf{Point-LIO} & \textbf{Ours} & \textbf{SR-LIVO} & \textbf{Fast-LIVO} & \textbf{Fast-LIVO2} & \textbf{Ours} \\
    \midrule
    escalator00   & 2.229 & 1.342 & - & - & 0.148 & 0.163 & 0.132  & 0.105 & \textbf{0.090} & 0.187 & 0.359 & 0.129 & \textbf{0.091} \\
    escalator01   & 0.225 & 0.300 & - & - & 0.143 & 0.152 & \ding{55} & 0.105 & \textbf{0.093} & 0.118 & 0.322 & 0.108 & \textbf{0.089} \\
    grass00       & \ding{55} & 0.789 & - & - & 0.115 & 0.407 & 0.210 & \textbf{0.092} & 0.101 & 0.103 & 0.563 & 0.231 & \textbf{0.096} \\
    room00        & 0.091 & 0.084 & - & - & 0.112 & 0.089 & 0.078 & 0.079 & \textbf{0.070} & 0.091 & 0.120 & 0.080 & \textbf{0.069} \\
    room01        & 0.095 & 0.093 & - & - & 0.120 & 0.098 & 0.095 & 0.097 & \textbf{0.084} & 0.106 & 0.120 & 0.095 & \textbf{0.083} \\
    underground00 & 0.759 & 1.091 & - & - & 0.769 & 0.551 & \textbf{0.381} & 0.439 & 0.748 & \textbf{0.385} & 0.894 & 1.342 & 0.832 \\
    \hline
    grass00       & \ding{55} & \ding{55} & - & - & 0.114 & 0.143 & 0.089 & \textbf{0.061} & 0.069 & 0.072 & 0.250 & 0.125 & \textbf{0.068} \\
    grass01       & 0.451 & \ding{55} & - & - & 0.762 & 0.182 & 0.208 & \ding{55} & \textbf{0.053} & \ding{55} & \ding{55} & 0.116 & \textbf{0.054} \\
    room00        & 0.075 & \ding{55} & - & - & 0.107 & 0.126 & 0.136 & 0.585 & \textbf{0.067} & \ding{55} & 0.175 & 0.193 & \textbf{0.066} \\
    transition00  & 0.070 & 0.085 & - & - & 0.115 & 0.086 & 0.078 & 0.077 & \textbf{0.051} & \ding{55} & 0.118 & 0.079 & \textbf{0.053} \\
    underground00 & 2.018 & 0.219 & - & - & 0.180 & 0.191 &0.203 & \textbf{0.093} & 0.168 & \textbf{0.084} & 0.444 & 0.186 & 0.144 \\
    \hline
    campus00      & 1.781 & 1.518 & 1.488 & \textbf{1.399} & 1.731 & 1.478 & \textbf{1.288} & 2.054 & 1.459 & \ding{55} & \textbf{1.395} & 1.640 & 1.488 \\
    parking00     & 0.277 & 0.289 & \textbf{0.282} & 0.301 & 0.353 & 0.486 & 0.529 & \textbf{0.246} & 0.303 & 0.286 & 0.661 & \textbf{0.239} & 0.300 \\
    parking01     & 0.313 & 0.534 & 0.411 & \textbf{0.354} & 0.474 & 0.789 & 2.065 & \textbf{0.318} & 0.332 & 0.347 & \ding{55} & 0.360 & \textbf{0.344} \\
    parking02     & 0.416 & 0.382 & 0.431 & \textbf{0.371} & 0.560 & 0.826 & 0.623 & 0.445 & \textbf{0.369}& 0.405 & 1.122 & 0.712 & \textbf{0.369} \\
    parking03     & 0.501 & 0.757 & 0.513 & \textbf{0.447} & 0.642 & 1.286 & 0.724 & 0.529 & \textbf{0.447} & 0.492 & 1.097 & 1.052 & \textbf{0.447} \\
    transition00  & 0.138 & 0.139 & - & - & 0.217 & 0.151 & 0.138 & \textbf{0.131} & 0.204 & \ding{55} & 0.173 & \textbf{0.146} & 0.150 \\
    transition01  & 0.208 & 0.189 & - & - & 0.255 & 0.203 & 0.185 & \textbf{0.173} & 0.196 & \ding{55} & 0.212 & 0.197 & \textbf{0.187} \\
    \hline
    \hline
    \end{tabular}
\end{table*}

\subsubsection{LiDAR-Camera Fusion}
For the LiDAR-camera configuration, we evaluate LXD-SLAM on the NTU-VIRAL dataset against SDV-LOAM~\cite{10086694}, a state-of-the-art (SOTA) LiDAR-visual odometry baseline. As reported in Table~\ref{tab:rmse-viral} (columns 2 and 3), our framework consistently outperforms SDV-LOAM across almost all sequences (except for the anomalous eee01), achieving a significant accuracy margin of at least 14~cm. 

This performance leap is primarily attributed to our robust {ray-to-mesh depth association} module. Unlike traditional visual-LiDAR frameworks that suffer from fragile feature-to-point associations or noisy depth projection, LXD-SLAM leverages the continuous, multi-layered GP surface model. This allows for highly accurate, ray-traced absolute depth recovery, which is sequentially refined via a 1D inverse depth filter, ensuring that only geometrically mature visual constraints participate in the joint IESKF update.

\subsubsection{LiDAR-LiDAR Fusion}
To verify our capability in processing multi-LiDAR configurations, we compare LXD-SLAM with MLOAM~\cite{jiao2021robust}, a representative baseline specifically optimized for multi-LiDAR setups. Despite an isolated divergence case in the challenging eee01 sequence, LXD-SLAM delivers superior trajectory tracking precision compared to MLOAM (Table~\ref{tab:rmse-viral}, columns 4 and 5). 

This superiority showcases the effectiveness of our cell-wise GP sub-mesh representation. By discretization and fitting up to three oriented surface layers per grid cell, our system natively pools unstructured multi-view point clouds into a unified, continuous geometric topology. This significantly enhances the descriptive power of the map and mitigates the geometric degeneracy or registration misalignment that typically plagues multi-sensor spatial aggregation.

\subsubsection{LiDAR-Inertial Fusion}
To assess LIO performance under aggressive and highly dynamic motions, we conduct benchmarks against a broad spectrum of recent competitive baselines, including D-LIOM~\cite{9760190}, Ct-LIO~\cite{10327749}, DLIO~\cite{dlio}, Point-LIO~\cite{bai2022pointlio}, LIO-SAM~\cite{shan2020lio}, and Fast-LIO2~\cite{xu2022fastlio2}. 

On the NTU-VIRAL dataset (Table~\ref{tab:rmse-viral}), LXD-SLAM secures the highest localization accuracy across all sequences, reducing the average ATE by approximately 4~cm compared to the runner-up method. Furthermore, as shown in Table~\ref{tab:rmse-fpv2} for the FusionPortableV2 dataset, our method dominates the majority of the sequences. Notably, LXD-SLAM exhibits a distinct advantage on handheld and quadruped robot platforms. This excellence stems directly from our {hierarchical prediction strategy} within the mathematically unified IESKF framework, where prioritized IMU propagation provides a highly reliable motion prior to distort point clouds, while the implicit point-to-mesh distance updates ensure stable convergence under rapid, erratic rotations.

\subsubsection{LiDAR-Inertial-Wheel Fusion}
For ground-moving robots, wheel encoders provide critical kinematic odometry constraints that are particularly vital in geometrically featureless or degraded scenarios. We validate this capability on the wheel-equipped sequences of FusionPortableV2, benchmarking against the SOTA LIWO~\cite{yuan2023liwo} framework. 

As summarized in Table~\ref{tab:rmse-fpv2}, LXD-SLAM yields higher localization precision in four out of the five test environments. More importantly, compared with the LiDAR-only configuration, the seamless injection of wheel encoder odometry markedly improves both the system's survival rate and structural accuracy. This underscores the merit of our adaptive state configuration, which effortlessly accommodates supplementary wheel odometry as a secondary dead-reckoning priority, safeguarding the estimation pipeline from structural degeneracy.

\subsubsection{LiDAR-Visual-Inertial Fusion}
Finally, we evaluate the complete tight-coupling capability under the LiDAR-Visual-Inertial (LCI) setup, comparing LXD-SLAM with leading multi-modal SOTA systems, including R$^3$LIVE~\cite{lin2022r3live}, SR-LIVO~\cite{srlivo}, Fast-LIVO~\cite{zheng2022fastlivo}, and Fast-LIVO2~\cite{fast-livo2}. 

The evaluation results across Tables~\ref{tab:rmse-viral} and~\ref{tab:rmse-fpv2} demonstrate that LXD-SLAM achieves top-tier tracking performance uniformly across all testing platforms—ranging from agile UAVs and rugged UGVs to quadruped robots and handheld devices. In terms of overall robustness on FusionPortableV2, our framework performs on par with Fast-LIVO2 and drastically outperforms Fast-LIVO and SR-LIVO. This compelling fusion efficacy stems from our mathematically consistent update step, which jointly minimizes primary point-to-mesh geometric errors and mature visual feature reprojection errors, fully unleashing the complementary strengths of geometric structures and visual textures.

\subsection{Results of Mapping}
\subsubsection{Global Consistent Dense Meshing}

\begin{figure*}
    \centering
    \includegraphics[width=\linewidth]{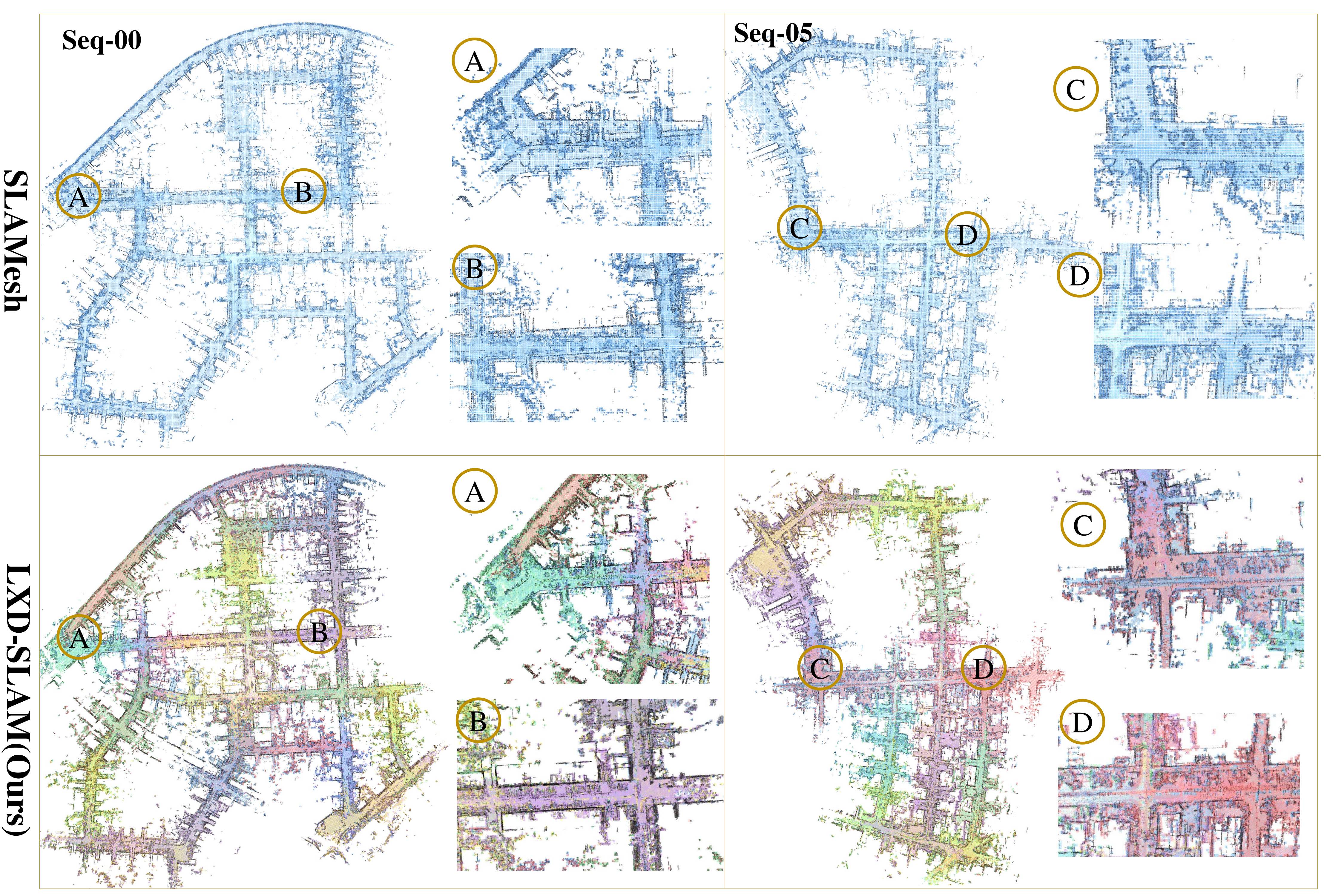}
    \caption{Globally Consistent Dense Mesh Construction in Large-Scale Scenes. Top: Dense mapping results of SLAMesh on KITTI Sequence-00 and Sequence-05. Significant map drifts (A/B/C/ D) are observed at loop closure junctions. Bottom: Globally consistent dense maps of LXD-SLAM on KITTI Sequence-00 and Sequence-05, demonstrating seamless alignment without any observable offsets at overlapping areas.}
    \label{fig:mesh-kitti}
\end{figure*}

In this section, we demonstrate the system's capability to construct geometrically consistent dense maps for large-scale outdoor environments. Fig. \ref{fig:mesh-kitti} shows the dense meshes corresponding to two large-scale sequences, Seq-00 and Seq-05 of KITTI Odometry. The trajectories of these sequences contain multiple loop closures; we mark the typical loop-closure locations in the figure and present the corresponding zoomed-in local views on their right. For SLAMesh~\cite{10161425}, since it represents the entire scene as a single map, we render its map in a uniform color. For LXD-SLAM, different sub-meshes are distinguished by different colors.

As can be observed, at loop-closure locations A and B in Seq-00, the sub-meshes reconstructed by our method overlap without noticeable misalignment, yielding a consistent dense reconstruction. In contrast, the mesh produced by SLAMesh exhibits evident offsets at the corresponding locations. Similarly, a consistent phenomenon can be observed at loop-closure locations C and D in Seq-05. These results convincingly demonstrate that the sub-mesh-based representation and loop-closure construction scheme of LXD-SLAM effectively addresses the challenge of consistent dense reconstruction in large-scale scenarios.

\subsubsection{Global Consistent Colorful Mapping}
\begin{figure*}
    \centering
    \includegraphics[width=0.95\linewidth]{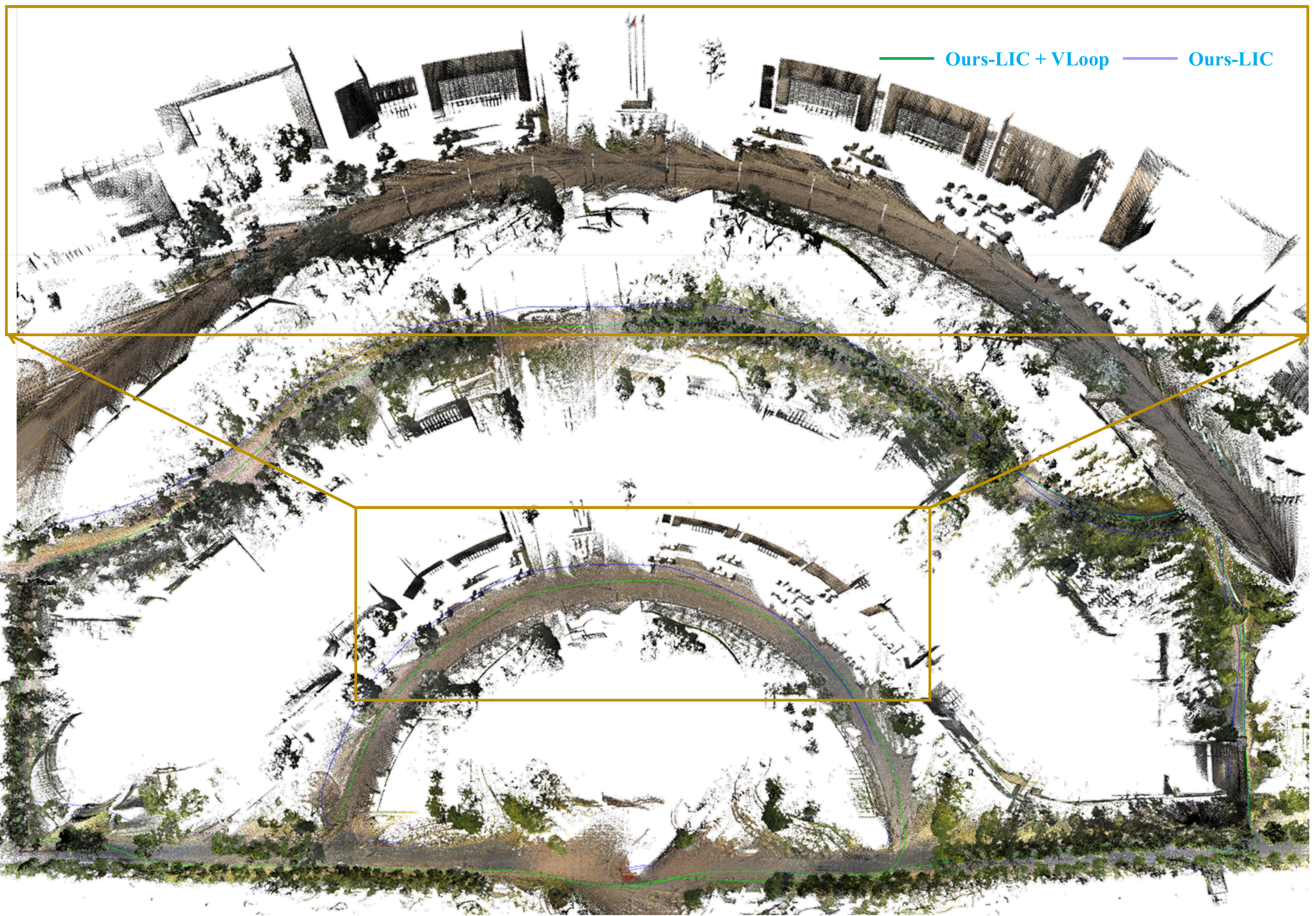}
    \caption{Globally consistent color mapping in campus environments. Leveraging backend loop closure detection and constraint construction, LXD‑SLAM builds a globally consistent color point cloud map for large‑scale scenes.}
    \label{fig:map-tj}
    \vspace{-3mm}
\end{figure*}

By leveraging back-end loop closure constraints and global pose graph optimization, our system not only generates dense meshes but also reconstructs globally consistent colorful point cloud maps. In a campus environment, we collected data over a large-scale scene using a custom-built handheld device. Fig. \ref{fig:map-tj} shows the consistent colored point cloud map reconstructed by LXD-SLAM. As can be seen, with the support of the loop closure detection and constraint construction, the full system that incorporates the back-end successfully eliminates accumulated drift (green trajectory), in contrast to the trajectory estimated by the pure odometry system (blue trajectory). The reconstructed map demonstrates excellent global consistency, and its accuracy can also be observed from the close-up views, where fine structures such as street lamps and flagpoles exhibit clearly discernible outlines and building walls are rendered with sharp edges and distinct corners.

\subsubsection{Drift-free Mapping with GNSS}
In the preceding subsections, we have validated the precise odometry and the consistent localization and reconstruction capabilities of LXD-SLAM in large-scale scenarios. However, when the system performs long-term mapping in large-scale environments without loop closures, GNSS information is virtually the only means to prevent the system from drifting. In this subsection, we present the drift-free localization and mapping results of LXD-SLAM with GNSS assistance.

\begin{figure*}
    \centering
    \includegraphics[width=0.95\linewidth]{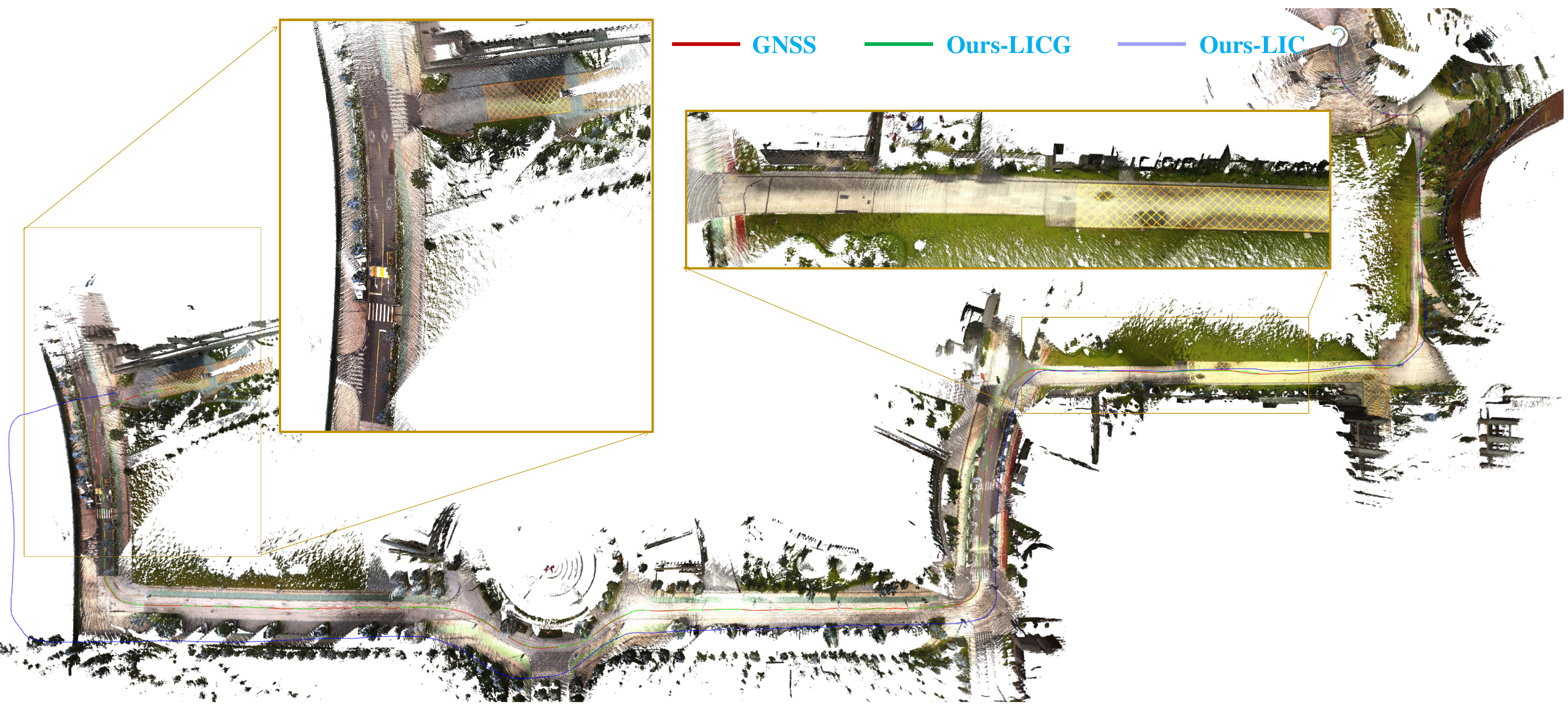}
    \caption{With the aid of GNSS, large-scale mapping with low drift is achieved. The fusion of GNSS signals in the backend enables the system to maintain both high accuracy in local mapping—as evidenced by the clear ground markings in magnified local views—and low drift over large-scale, long-duration mapping sessions even in the absence of loop closures, which is corroborated by the high consistency between the estimated trajectory and the GNSS reference trajectory.}
    \label{fig:map-gnss}
    \vspace{-3mm}
\end{figure*}

As illustrated in Fig. \ref{fig:map-gnss}, we plot the trajectory of LXD-SLAM (LICG), the trajectory of LXD-SLAM (LIC), the GNSS trajectory, and the corresponding colored point cloud under LICG poses. It can be seen from the trajectories that LXD-SLAM (LICG) nearly overlaps with the GNSS trajectory, indicating that the fusion of GNSS information yields low-drift long-term localization. In contrast, LXD-SLAM (LIC) exhibits pronounced drift, which is difficult to eliminate in the absence of GNSS signals. From the zoomed-in view at the endpoint, the map generated by our method demonstrates high accuracy, with road markings such as zebra crossings and grid lines clearly distinguishable.

\subsection{Run Time}

\begin{table*}[htbp]
\centering
\caption{Mean time costs of key modules of LXD-SLAM on different scenarios (ms). LR-Guess/VR-Guess means the LiDAR-based/Visual-based rotation guess. L-Update/V-Update denotes the LiDAR/Visual Kalman update step. L/LL/LC/LI/LCI is the abbreviation of LiDAR/LiDAR-LiDAR/LiDAR-Camera/LiDAR-IMU/LiDAR-Camera-IMU, respectively.}
\label{tab:runtime}
\begin{tabular}{c|cccccc|ccccc}

\hline
\hline
~ & \multicolumn{6}{c|}{\textbf{Key Modules}} & \multicolumn{5}{c}{\textbf{Odometry}} \\ \cline{2-12}
\textbf{Scenario} & \textbf{LR-Guess} & \textbf{VR-Guess} & \textbf{Meshing} & \textbf{Ray-Casting} & \textbf{L-Update} & \textbf{V-Update} & \textbf{L} & \textbf{LL} & \textbf{LC} & \textbf{LI} & \textbf{LCI} \\ \hline
\textbf{eee } & 5.33 & 7.68 & 22.00 & 5.84 & 7.03 & 1.70 & 35.50 & 37.10 & 57.23 & 33.84 & 52.95 \\ 
\textbf{nya } & 4.85 & 8.10 & 17.88 & 5.67 & 7.09 & 2.10 & 31.57 & 32.62 & 56.67 & 30.73 & 48.54 \\ 
\textbf{sbs } & 5.04 & 7.61 & 16.96 & 5.61 & 6.47 & 1.71 & 33.19 & 34.59 & 53.67 & 27.59 & 47.96 \\
\hline
\hline
\end{tabular}
\end{table*}

\subsubsection{Time costs of key modules}
As shown in Table~\ref{tab:runtime}, the most time-consuming modules in the system are meshing and LiDAR update, followed by initial rotation estimation (LR-Guess and VR-Guess) and ray-casting, while visual update takes the least time. This phenomenon is readily understandable: dense meshing of the map requires performing Gaussian process surface updates on the cells in all overlapping regions. Since the system's pose estimation relies primarily on the point cloud first, with visual update performed subsequently, LiDAR update accounts for the bulk of the update time. Moreover, although the prediction step consumes a considerable amount of time, we will demonstrate later that this module is indispensable for enhancing the system's robustness under rapid rotation.

\subsubsection{Time costs of different combinations}
Regarding the combination of different modules, we observe that incorporating multiple LiDARs does not significantly increase the time consumption compared with using a single LiDAR. The primary reason is that, apart from the preprocessing synchronization step, no special processing is applied to multi-LiDAR data. The additional time incurred mainly arises from the dense reconstruction of point clouds from new viewpoints, which can be substantially mitigated through parallelization and therefore does not lead to a notable increase in time. Another prominent observation is that the LI combination achieves the highest efficiency among all configurations. This is largely because the introduction of the IMU provides a good initial estimate for the system, thereby reducing the time required for pose updates. Additionally, it is also evident that incorporating visual information introduces a considerable increase in time consumption. Although we adopt sparse feature points to reduce the computational load, the processes of ray-casting and feature point maintenance still incur necessary additional time.

\subsection{Degeneration Case}
\begin{figure}
    \centering
    \includegraphics[width=0.9\linewidth]{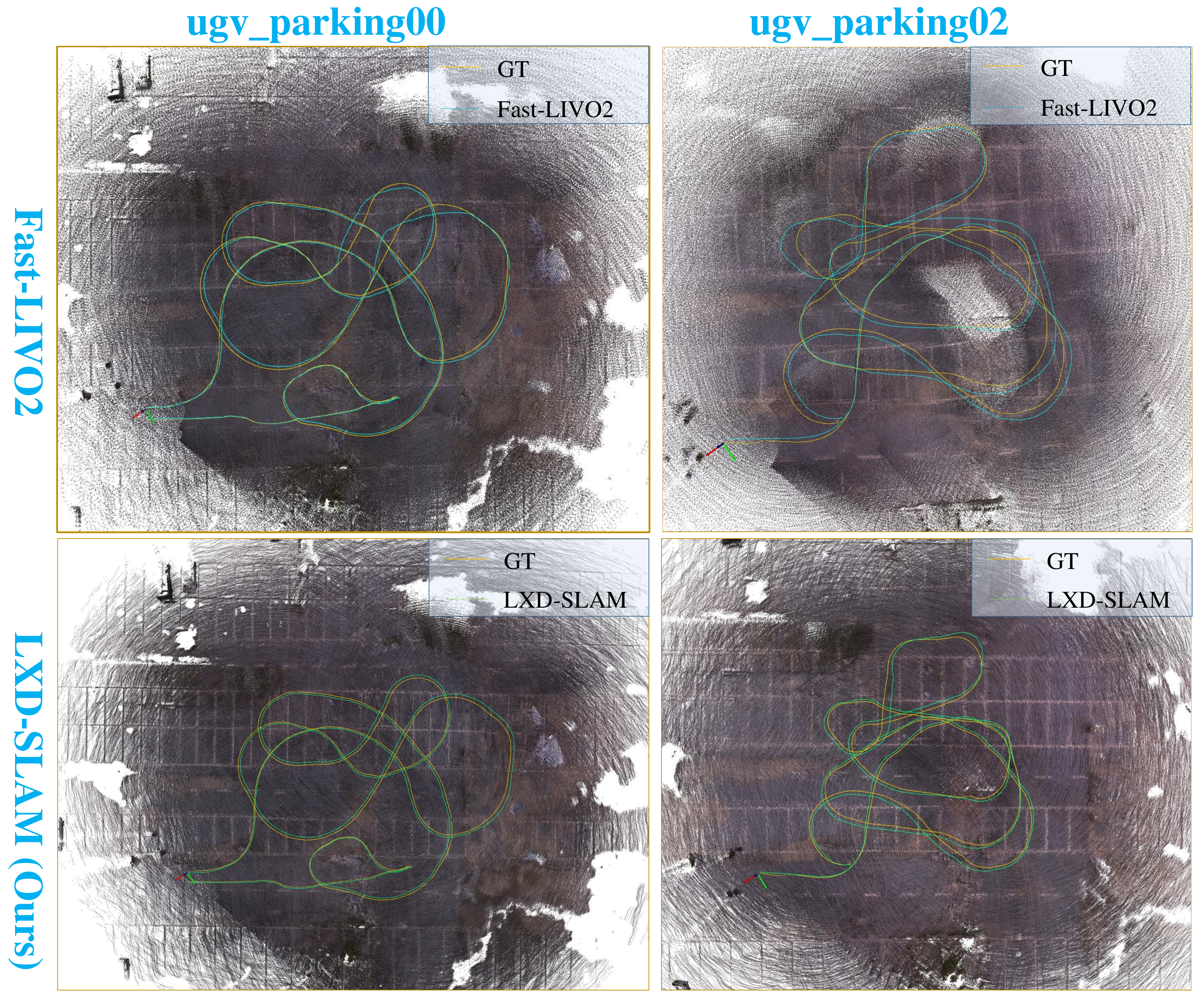}
    \caption{Resulting maps under degeneration. Top: maps and trajectories produced by Fast-LIVO2. Bottom: maps and trajectories estimated by our LXD-SLAM.}
    \label{fig:map-degenerate}
\end{figure}

One of the objectives of multi-sensor fusion is to enhance the localization and mapping accuracy of the system, which has been corroborated in the preceding sections. Another primary purpose is to improve the system's robustness in complex environments or under challenging motions. To illustrate this, in this section, we validate the system's robustness in degenerate environments on the challenging parking\_00 and parking\_02 sequences from the FusionPortableV2 dataset. The parking scenario itself is relatively open; to further simulate degeneracy, we artificially retain only the point cloud within a range of 30 meters from the vehicle body. As a result, the system lacks sufficient LiDAR constraints in the horizontal direction and must rely on visual information for compensation.

We evaluate both Fast-LIVO2 and LXD-SLAM (LCI) under such challenging case. The estimated trajectories, ground-truth trajectories, and the corresponding colored point cloud maps are plotted in Fig. \ref{fig:map-degenerate}. As can be observed, both parking\_00 and parking\_02 involve complex rotational motions, posing significant challenges for pose estimation. Nevertheless, both our LXD-SLAM and Fast-LIVO2 achieve successful localization in this scenario. In terms of mapping accuracy, compared with Fast-LIVO2, the map reconstructed by our method is sharper and exhibits more accurate parking lane markings. This indirectly demonstrates the robustness of our system in handling challenging degenerate environments.

\subsection{Ablation Study}
\subsubsection{Rotation Guess}
\begin{table}[htp]
    \centering
    \caption{Ablation studies on rotation guess.}
    \begin{tabular}{c|cc|cc}
    \hline
    \hline
    ~ & w/o LR & with LR & w/o VR & with VR\\
    \hline
         eee\_01& 10.097&\textbf{0.147} & 5.657 & \textbf{1.952} \\
         eee\_02& 2.219& \textbf{0.071}&  4.827&  \textbf{0.066}\\
         eee\_03& 2.030& \textbf{0.237}& 0.446& \textbf{0.102}\\
         nya\_01& 0.404& \textbf{0.071} & 0.380&\textbf{0.054} \\
         nya\_02& \textbf{0.088}& 0.105& 0.201&  \textbf{0.073}\\
         nya\_03& \textbf{0.070}& 0.078& \textbf{0.070} &  \textbf{0.070}\\
         sbs\_01& \textbf{0.068}& 0.070& 0.086&  \textbf{0.062}\\
         sbs\_02& \textbf{0.130}& \textbf{0.130}&  0.135& \textbf{0.072} \\
         sbs\_03& 0.080& \textbf{0.073}& 2.021& \textbf{0.060} \\
    \hline
    \hline
    \end{tabular}
    \label{tab:ablation}
\end{table}

As shown in Table~\ref{tab:ablation}, when the LiDAR-based rotation guess is enabled, the system exhibits significantly improved localization stability, especially in the eee scenarios. In terms of localization accuracy, although a slight degradation is observed in a few cases with LiDAR Rotation guess (LR) enabled, noticeable accuracy gains are achieved in the majority of cases. Similarly, when the Vision-based Rotation guess (VR) is enabled, the system also demonstrates markedly enhanced localization stability in the eee scenarios. Regarding localization accuracy, with the aid of visual guess, the system achieves a substantial improvement in accuracy compared with the LC fusion system without guess. These observations collectively indicate that both LiDAR-based and vision-based rotation guess offer significant positive benefits for pure LiDAR or LiDAR–camera sensor configurations.

\begin{figure}
    \centering
    \includegraphics[width=0.95\linewidth]{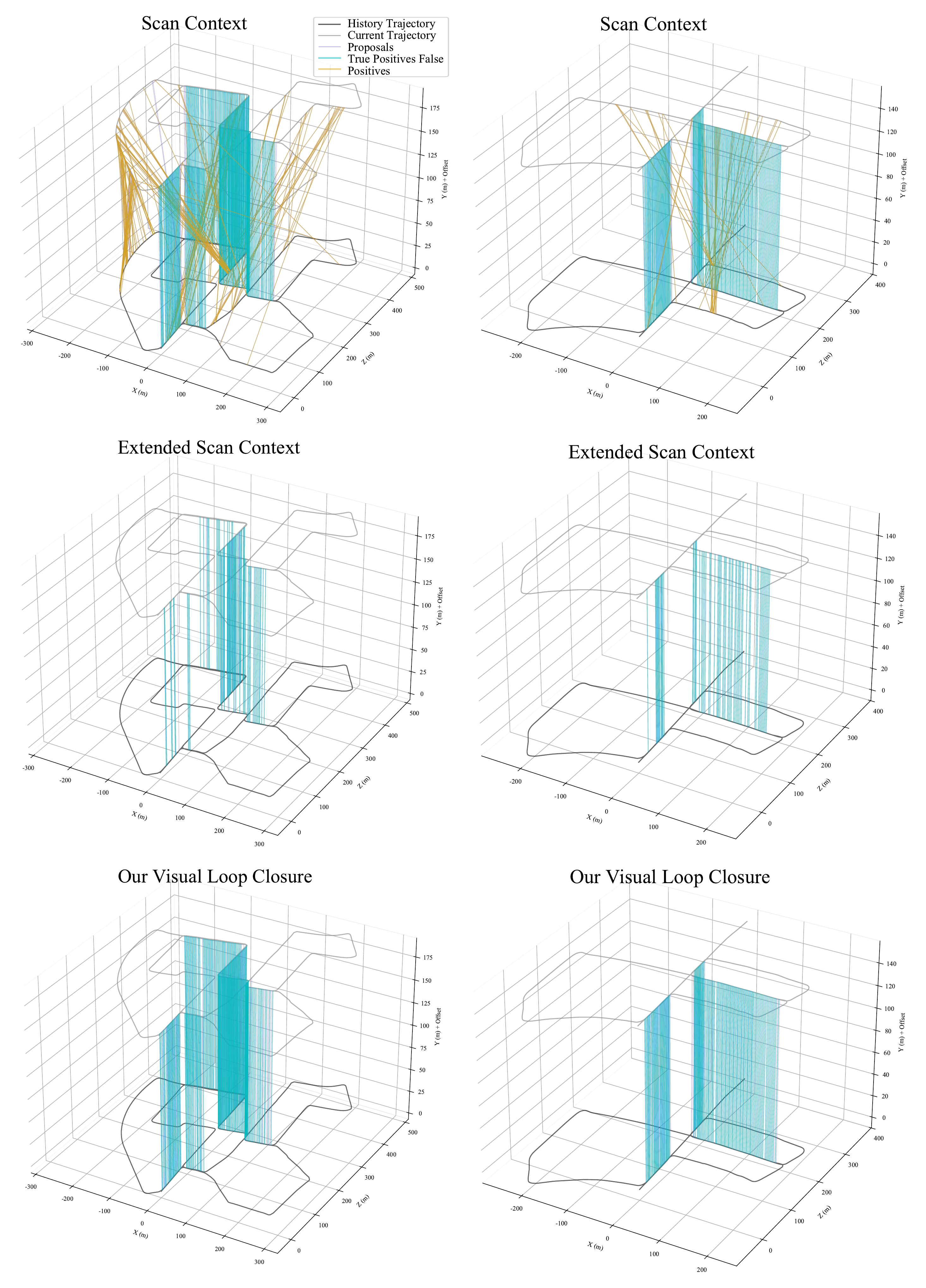}
    \caption{Ablation studies on the extended scan context and visual loop closure construction stategy.}
    \label{fig:loops}
\end{figure}

\subsubsection{Extended Scan Context and Bi-directional PnP}

While Scan Context~\cite{scancontextpp} has virtually become the gold standard for loop closure detection with full-perspective point clouds, its performance degrades significantly when the LiDAR field of view (FoV) is restricted. To simulate such scenarios, we retain only the frontal portion of the KITTI point clouds (i.e., points where $x > 0$). Due to the packaging constraints in vehicle design, this restricted-FoV condition is, in fact, quite prevalent in autonomous driving systems. 

The detected loop closures and the successfully established constraint pairs are visualized in Fig. \ref{fig:loops}, where light red line segments indicate correct loop candidates, orange segments represent false positive detections, and green segments denote loop constraints with accurately established 6-DoF poses. As can be observed, under a restricted LiDAR FoV, Scan Context generates a substantial number of false loop closures, which would be fatal to the system's consistent localization and mapping. In contrast, although our extended scan context exhibits a slightly lower recall rate, it produces virtually no false positives and simultaneously guarantees the reliable formulation of correct loop constraints. Under these circumstances, the visual loop closure module generates a greater number of loop candidates and establishes more correct loop constraints, thereby ensuring the timely elimination of pose drift and the construction of globally consistent dense maps.


\section{Conclusion}
In this paper, we presented LXD-SLAM, a unified, modular, and mathematically consistent multi-sensor fusion dense SLAM framework with LiDAR at its core. By deriving the sensor configuration from the power set of five modalities—LiDAR, Camera, IMU, Wheel Encoder, and GNSS—the proposed system successfully achieves true plug-and-play versatility, supporting up to $2^5$ distinct sensor combinations without requiring structural re-engineering.
The excellence of LXD-SLAM is underpinned by three core methodological components: a) An Iterative Kalman Filter featuring a hierarchical prediction strategy and unified point-to-mesh/reprojection update signals.
b) A robust visual feature manager that efficiently tracks features and recovers absolute depths via ray tracing against a continuous, multi-layered Gaussian Process (GP) surface representation.
c) A hybrid factor graph optimization integrated with an Extended Scan Context (ESC) descriptor and Bidirectional PnP visual loop closure for drift-free trajectory and map refinement.
Extensive evaluations on public datasets and real-world experiments demonstrate that LXD-SLAM either matches or outperforms specialized state-of-the-art solutions across various configuration modes. More importantly, the system's capability to construct high-fidelity, globally consistent dense meshes in real-time offers a versatile and robust perception foundation for diverse downstream autonomous robotic tasks.

\normalem
\bibliographystyle{ieeetr}
\bibliography{ref}

\end{document}